\pdfoutput=1
\documentclass[acmsmall,noacm]{acmart}
\AtBeginDocument{%
  }

\setcopyright{acmlicensed}
\copyrightyear{2025}
\acmYear{2025}
\acmDOI{10.1145/3789261}
\usepackage{graphicx}
\usepackage{array}
\usepackage{tabularx}
\usepackage{tcolorbox}     
\usepackage{fontawesome5}  
\usepackage{hyperref}      

\usepackage{multirow} 
\usepackage{pifont}
\usepackage{tikz}
\usepackage[edges]{forest}  
\usepackage{hyperref}
\hypersetup{
    colorlinks=true,
    linkcolor=blue, 
    filecolor=blue,
    urlcolor=blue,
    citecolor=blue
}
\acmVolume{58}
\acmNumber{9}
\acmArticle{223}
\acmMonth{2}

\usepackage{tikz}
\usetikzlibrary{positioning,arrows.meta,calc}
\usepackage{hyperref}      
\usepackage{fontawesome5}  

\definecolor{health}{RGB}{169,194,235}
\definecolor{sci}{RGB}{210,210,210}
\definecolor{emb}{RGB}{190,220,190}
\definecolor{fin}{RGB}{240,200,180}
\definecolor{code}{RGB}{250,230,130}

\tikzset{
  >={Stealth[length=3mm,width=2mm]},
  every node/.style = {font=\small},
  cat/.style  = {draw, rounded corners=3pt, fill=#1,
                 inner sep=4pt, text width=4cm, align=left},
  work/.style = {draw, rounded corners=3pt, fill=#1!60!white,
                 inner sep=4pt, text width=8cm,   align=left,
                 anchor=west},
  root/.style = {draw, rounded corners=3pt, fill=white,
                 inner sep=5pt, font=\bfseries\large, align=center}
}

\begin{document}

\title{A Survey on the Optimization of Large Language Model-based Agents}

\author{Shangheng Du}

\orcid{0000-0002-5792-631X}
\affiliation{
  \institution{Shanghai Institute of Artificial Intelligence for Education, East China Normal University; School of Computer Science and Technology, East China Normal University}
  \state{Shanghai}
  \country{China}
}
\email{dsh@stu.ecnu.edu.cn}

\author{Jiabao Zhao}
\orcid{0000-0002-0691-5741}
\authornote{Corresponding author.}
\affiliation{%
  \institution{School of Computer Science and Technology, Donghua University}
  \state{Shanghai}
  \country{China}
}
\email{jbzhao@dhu.edu.cn}

\author{Jinxin Shi}
\orcid{0009-0009-8898-6030}
\affiliation{
  \institution{Shanghai Institute of Artificial Intelligence for Education, East China Normal University; School of Computer Science and Technology, East China Normal University}
  \state{Shanghai}
  \country{China}
}
\email{jinxinshi@stu.ecnu.edu.cn}

\author{Zhentao Xie}
\author{Xin Jiang}
\author{Yanhong Bai}
\author{Liang He}
\orcid{0000-0002-4723-5486}
\affiliation{
  \institution{School of Computer Science and Technology, East China Normal University}
  \state{Shanghai}
  \country{China}
}
\email{ecnudavidtao@gmail.com}
\email{51275901099@stu.ecnu.edu.cn}
\email{Lucky_Baiyh@stu.ecnu.edu.cn}
\email{lhe@cs.ecnu.edu.cn}

\thanks{This work is supported by the National Natural Science Foundation of China (Grant Nos. 62477011, 62207013).
}
\renewcommand{\shortauthors}{S. Du et al.}

\begin{abstract}
With the rapid development of Large Language Models (LLMs), LLM-based agents have been widely adopted in various fields, becoming essential for autonomous decision-making and interactive tasks. However, current work typically relies on prompt design or fine-tuning strategies applied to vanilla LLMs, which often leads to limited effectiveness in complex agent-related environments.
Although numerous recent studies have explored various strategies to optimize LLM-based agents for complex agent tasks, a systematic review summarizing and comparing these methods from a holistic perspective remains lacking. In this survey, we provide a comprehensive review of LLM-based agent optimization approaches, categorizing them into parameter-driven and parameter-free methods. We first focus on parameter-driven optimization, covering fine-tuning-based optimization, reinforcement learning-based optimization, and hybrid strategies, analyzing key aspects such as trajectory data construction, reward function design, and optimization algorithms. Additionally, we briefly discuss parameter-free strategies that optimize agent behavior through prompt engineering and external knowledge retrieval. Finally, we summarize the evaluation for agents, review key applications of LLM-based agents, and discuss the major challenges and promising future directions. A curated collection of the surveyed works is provided at \href{https://github.com/YoungDubbyDu/LLM-Agent-Optimization}{https://github.com/YoungDubbyDu/LLM-Agent-Optimization}.

\end{abstract}


\received{20 April 2025}
\received[revised]{7 November 2025}
\received[accepted]{5 June 2009}

\maketitle
\section{Introduction}

The development of autonomous agents has been a long-term pursuit in Artificial Intelligence (AI). AI agents have evolved from early rule-based and expert system-based architectures to reinforcement learning (RL)-driven agents, which are now widely applied in many fields~\cite{rl}. Traditional RL-based agents optimize policies through interaction with environments, using structured reward functions to achieve goals and improve performance over time. However, these approaches often require extensive training, rely on well-defined state and action spaces, and struggle with generalization across diverse tasks.

In recent years, Large Language Models (LLMs) such as GPT-5~\cite{gpt4} and Deepseek-r1~\cite{deepseek} have achieved remarkable success, demonstrating exceptional capabilities in language understanding, reasoning, planning and complex decision-making. Building on these strengths, LLMs can serve as agents, providing a promising pathway to improve autonomous decision-making and achieve AGI~\cite{riseagent}. Unlike conventional RL-based agents, which optimize explicit reward-driven policies, LLM-based agents operate through text-based instructions and prompt templates and in-context learning (ICL), allowing greater flexibility and generalization. These agents leverage the comprehension and reasoning capabilities of LLMs to interact with environments through natural language, execute complex multi-step tasks, and dynamically adapt to evolving scenarios. Existing LLM-based agents utilize various methods such as task decomposition~\cite{agentplan}, self-reflection~\cite{selfreflection}, memory augmentation~\cite{memsurvey}, and multi-agent collaboration~\cite{masurvey} to achieve high performance in a range of domains, including software development~\cite{software}, embodied intelligence~\cite{epo}, web navigation~\cite{mind2web}.

However, despite their strengths, LLMs are not inherently designed for autonomous decision-making or long-term tasks. Their training objectives focus on next-token prediction rather than the reasoning, planning, or interactive learning required for agent-based behaviors. This distinction also separates LLM-based agents from tool-augmented reasoning models: While tool-augmented LLMs can call external functions, they remain fundamentally reactive, executing tools only when prompted. In contrast, LLM-based agents employ the model as a cognitive core within an agentic workflow, enabling the system to initiate actions, plan strategically, select tools autonomously, and regulate its own behavior through reflection. As a result, deploying LLMs as agents in complex environments presents several key challenges: (1) long-horizon planning and multi-step reasoning often degrade over extended interactions; (2) limited memory prevents effective use of past experience; and (3) reliance on static pre-training reduces adaptability to novel or dynamic environments. Such issues are especially pronounced in open-source LLMs, which still trail proprietary models like GPT-5 in agentic capability. At the same time, the high cost and opacity of closed-source systems further highlight the need for optimizing open LLMs for agent performance.

Existing techniques, such as supervised fine-tuning (SFT)~\cite{ftsurvey} and reinforcement learning with human feedback (RLHF)~\cite{rlhf}, have made significant strides in improving LLM performance in instruction following tasks, but they fail to fully address the challenges of decision-making, long-term planning, and adaptability for LLM-based agents. 
Optimizing LLM-based agents requires a deeper understanding of dynamic environments and agent behaviors, as well as the design of more suitable algorithms tailored to agent-specific demands.
To address these challenges, recent studies have explored various optimization strategies that help LLM-based agents refine their decisions, self-evolve, and handle dynamic, complex tasks.

In this paper, we provide a comprehensive survey on LLM-based agent optimization, systematically categorizing methods into parameter-driven and parameter-free optimization strategies. This survey focuses on methodologies for enhancing LLM-based agent capabilities through tuning, RL, and other optimization strategies.
Specifically, \textbf{Parameter-driven Optimization} refers to adjusting LLM parameters to enhance agent performance. We categorize existing methods into three types: (1) conventional fine-tuning-based optimization, consisting of two key stages: agent trajectory construction and agent fine-tuning; (2) RL-based optimization, further divided into reward function-based methods leveraging traditional RL algorithms like Actor-Critic~\cite{actor-critic} and Proximal Policy Optimization (PPO) \cite{ppo}, and preference alignment-based methods utilizing Direct Preference Optimization (DPO) \cite{DPO} to align agent policies with human preferences or task objectives; (3) hybrid fine-tuning optimization, an emerging direction that integrates SFT with RL to iteratively refine agent behaviors.
In addition, we outline \textbf{Parameter-free Optimization} methods, which aim to improve agent behavior without modifying model parameters. These methods leverage prompt engineering, in-context learning and retrieval-augmented generation (RAG), incorporating various types of information into prompts to guide agents' actions. We categorize them into experience-based, feedback-based, tool-based, Retrieval-based, and multi-agent collaborative optimization. 
To ensure reproducibility and facilitate practical use, we provide an extended list with representative agent optimization methods (including paper links and open-source code repositories) at \href{https://github.com/YoungDubbyDu/LLM-Agent-Optimization}{https://github.com/YoungDubbyDu/LLM-Agent-Optimization}.

\begin{figure}[htbp]
\setlength{\abovecaptionskip}{0.1cm}
\centerline{\includegraphics[width=\linewidth]{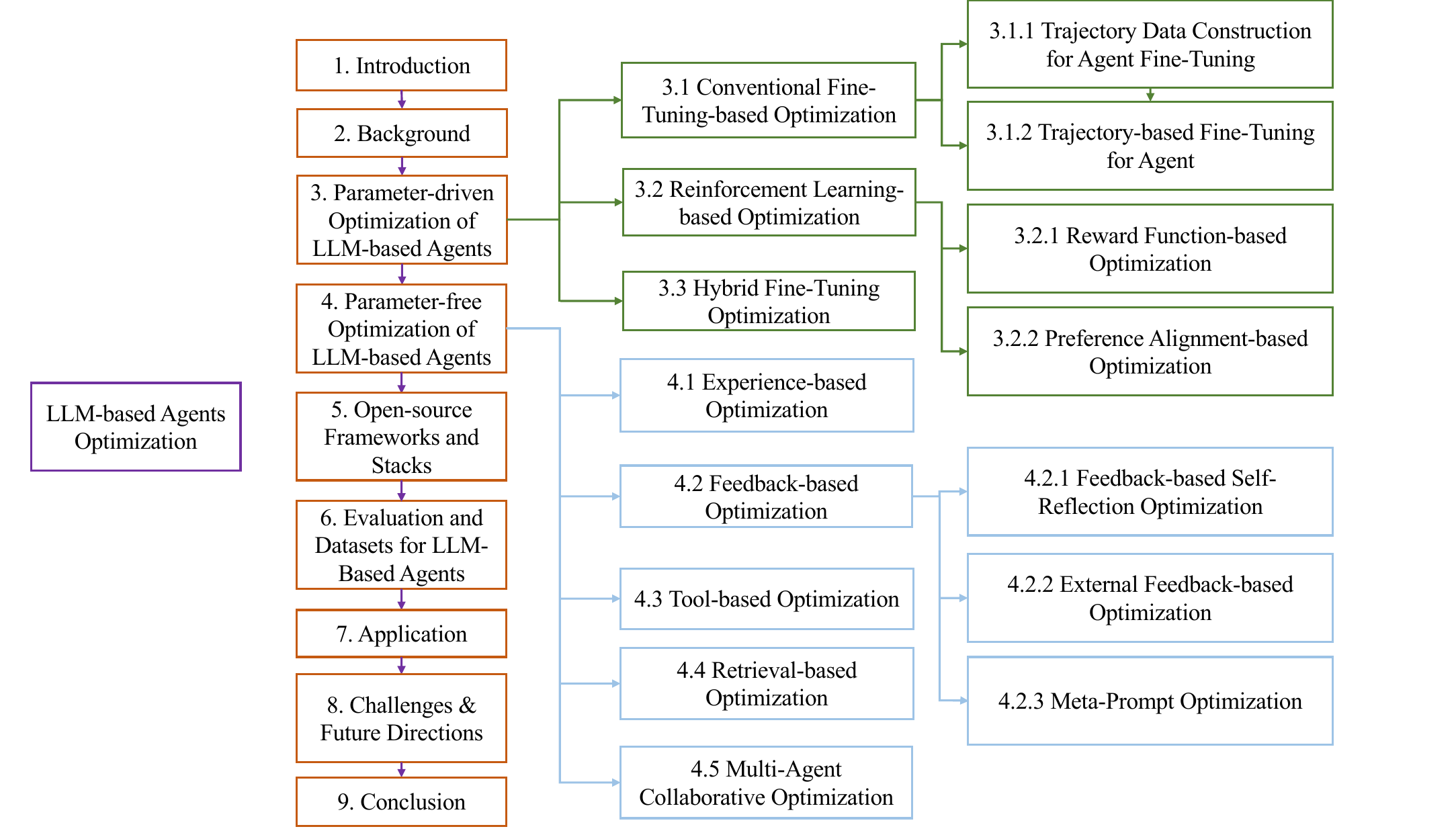}}
\caption{An Overview of the Paper Organization.}
\label{paper organization}
\end{figure}

\textbf{Comparison to related surveys.} While LLM-based agents have attracted increasing attention, existing surveys primarily focus on general LLM optimization~\cite{ftsurvey,lorasurvey,selfevolutionsurvey} or specific agent components such as planning~\cite{agentplan}, memory~\cite{memsurvey}, and multi-agent coordination~\cite{masurvey}. However, they do not treat LLM-based agent optimization as a distinct research topic. In contrast, this work is the first survey on LLM-based agent optimization techniques, facilitating a clearer understanding and comparison of existing methods and providing directions for future research.

\textbf{Scope and rationales.} (1) We survey only LLM-based agent optimization methods aimed at enhancing agent capabilities such as problem-solving and task execution, covering both parameter-driven and parameter-free approaches. We exclude works centered on general LLM efficiency, role-playing, or dialogue; (2) Our selection includes papers from top AI and NLP conferences and journals, as well as recent high-impact arXiv preprints to capture the latest developments. (3) We focus on studies since 2022 to reflect recent advancements in LLM-based agent optimization.

\textbf{Our key contributions} can be summarized as follows:
\begin{itemize}
    \item We present a comprehensive survey focused on the optimization of LLM-based agents. To the best of our knowledge, this is the first systematic review dedicated to this line of research.
    \item We categorize optimization methods by their underlying techniques and analyze them in depth from both algorithmic and workflow perspectives, covering parameter-driven and parameter-free approaches.
    \item We summarize widely used open-source frameworks, evaluation, datasets, highlight representative applications across domains, and discuss current challenges and future directions.
\end{itemize}

\textbf{Organization of this survey.} The schematic layout of this manuscript is illustrated in Figure \ref{paper organization}. Section \ref{sec2} provides the background knowledge and related concepts. In Section \ref{sec3}, we systematically review parameter-driven optimization approaches. Section \ref{sec4} classifies and summarizes parameter-free optimization methods.
Section~\ref{sec5} introduces open-source agent frameworks and stacks. Section~\ref{sec6} presents datasets and evaluation benchmarks, while Section \ref{sec7} reviews agent practical applications in various domains. Finally, Section \ref{sec8} highlights challenges and future directions.

\section{Background}\label{sec2}
\subsection{Reinforcement Learning-based Agent Optimization}
RL has long provided a foundation for agent optimization by enabling agents to learn from interactions with environments. Existing approaches primarily fall into value-based and policy-based methods~\cite{rl,value_policy,rloverview}. Value-based algorithms such as Q-learning~\cite{qlearning,qlsurvey} estimate action values to guide decision-making, while policy-based algorithms, including Policy Gradient and PPO~\cite{pg,pgview,ppo}, directly optimize policies through reward gradients, often improving stability via constrained updates. Actor–Critic methods~\cite{actor-critic} integrate value estimation with policy learning to enhance convergence. Beyond single-agent settings, Multi-Agent RL (MARL) extends these principles to cooperative and competitive multi-agent interaction~\cite{marl1,marl2}.

Recent developments increasingly focus on aligning agent behavior with human preferences. RLHF~\cite{rlhf} incorporates human feedback into policy optimization, and DPO~\cite{DPO} further streamlines preference-based learning by removing explicit reward modeling. Overall, RL-based optimization has progressed from classical value and policy learning to preference-driven and multi-agent formulations, offering essential foundations for decision-making in LLM-based agents.

\subsection{LLM Fine-Tuning}
LLM fine-tuning is a key technique for adapting pre-trained models to specific tasks through optimizing parameters. The most widely used approach is SFT, where LLMs are trained on labeled data to improve task-specific performance.
Instruction tuning, a common variant of SFT, enhances LLMs’ ability to follow human instructions by training on instruction–response pairs~\cite{instructionsurvey, visualit}. Another major development is parameter-efficient fine-tuning (PEFT), which updates only a small subset of parameters to significantly reduce computational cost while preserving performance. Representative methods include P-Tuning~\cite{ptuning}, LoRA~\cite{lora}, and QLoRA~\cite{qlora}. 
Additionally, RLHF is used to fine-tune LLMs by integrating human feedback, improving their decisions and aligning outputs with user preferences~\cite{rlhf}.
These techniques enable LLMs to adapt more efficiently to a wide range of tasks, enhancing their effectiveness in real-world scenarios.

\section{Parameter-driven Optimization of LLM-based Agents}\label{sec3}

\textbf{Comparison with LLM parameter optimization.} Parameter-driven LLM optimization focuses on "how to create a better model", aiming to enhance general language understanding, instruction following, and broad task performance. In contrast, LLM-based agent parameter optimization addresses "how to better use the model to solve complex agent tasks", emphasizing task-specific capabilities such as decision-making and multi-step planning in dynamic environments.

 In this section, we discuss how parameter-driven optimization methods improve the performance of LLM-based agents. Specifically, we categorize these methods into three main technical approaches according to different strategies for parameter tuning: conventional fine-tuning-based optimization, RL-based optimization, and hybrid optimization. 

\subsection{Conventional Fine-Tuning-based Optimization}\label{sec3.1}
\vspace{-0.5em}
\begin{figure}[h]
\setlength{\abovecaptionskip}{0.1cm}
\centerline{\includegraphics[width=0.9\linewidth]{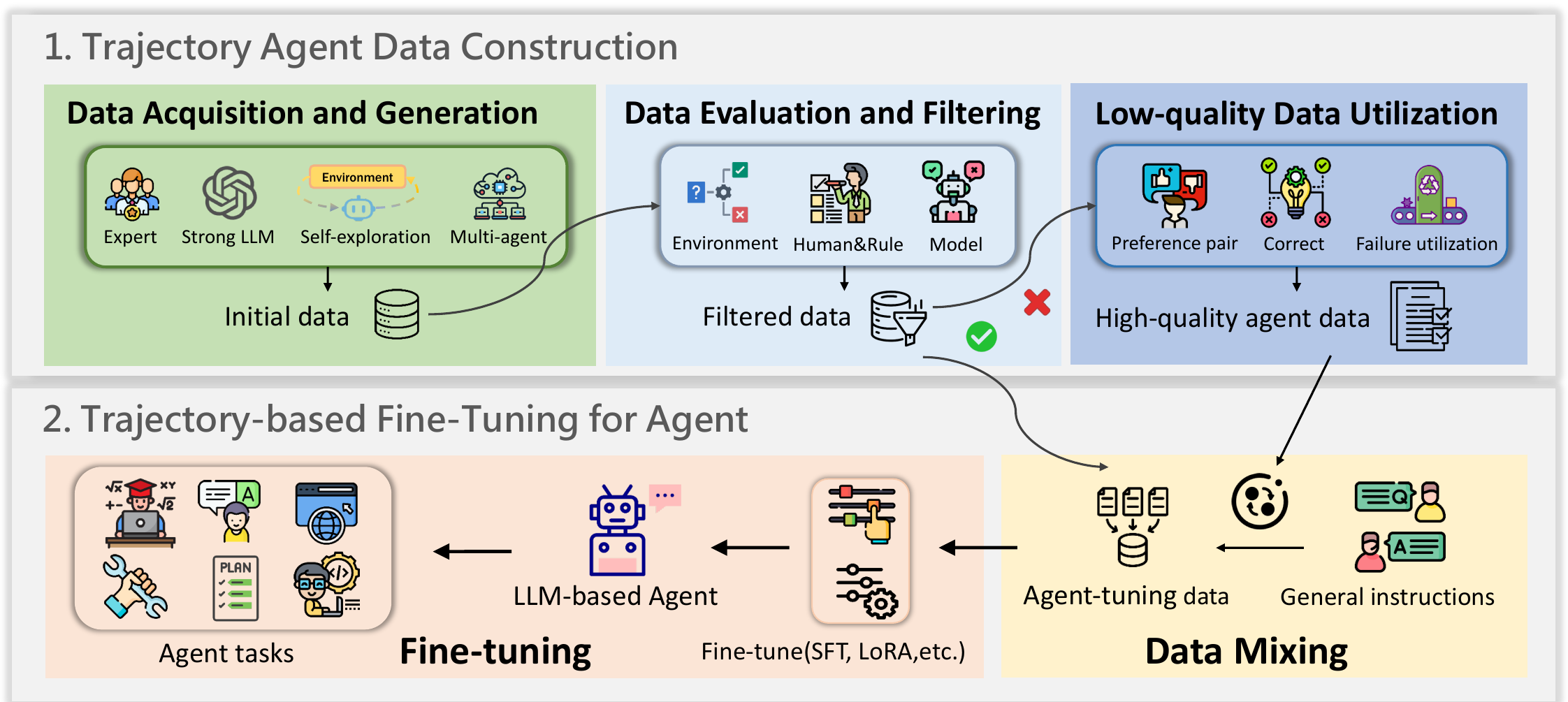}}
\caption{Workflow of Fine-Tuning-based Optimization for LLM-based Agents.}
\label{agenttuning}
\end{figure}

Conventional fine-tuning–based agent optimization adapts LLM parameters through instruction tuning and PEFT methods, using trajectory datasets typically formatted as SFT to align models with task-specific behaviors. This process generally involves two components: (1) constructing high-quality trajectories tailored to agent tasks, and (2) fine-tuning the LLM-based agent (workflow shown in Figure~\ref{agenttuning}). Methods involving RL-based optimization (e.g., PPO, DPO) are discussed separately in §\ref{sec3.2}. Here, we focus on traditional fine-tuning techniques employed in existing agent-optimization systems, outlining how each stage contributes to improving agent capabilities. Table~\ref{ft_optimization_tab} summarizes representative approaches, highlighting data preparation and fine-tuning strategies.


\begin{table}[htbp]
\setlength{\abovecaptionskip}{0.1cm}
\renewcommand{\arraystretch}{1}
\caption{Comparison of Conventional Fine-Tuning-based Optimization for LLM-based Agents: Data Construction and Fine-Tuning. Note: MA - Multi-Agent Framework; LQ - Low-Quality Data Utilization.}
\begin{center}
\resizebox{\textwidth}{!}{%
\begin{tabular}{c|cccc|cc}  
\hline
\multirow{2}{*}{\textbf{Method}} & \multicolumn{4}{c|}{\textbf{Trajectory Agent Data Construction}} & \multicolumn{2}{c}{\textbf{Fine-Tuning}} \\
\cline{2-7}
 & \textbf{Generation} & \textbf{Filtering} & \textbf{MA} & \textbf{LQ} & \textbf{Fine-tune Approach} & \textbf{Base Model} \\
\hline
AgentTuning~\cite{agenttuning} & Strong LLM & Human or Rule & \checkmark & \checkmark & Instruction Tuning & Llama-2-7B/13B/70B \\
\hline
SMART~\cite{smart} & Multi-agent & Environment & / & \checkmark & LoRA & Llama-2-7B \\
\hline
Agent-FLAN~\cite{agentflan}& Expert & Model & \checkmark & \checkmark & Instruction Tuning & Llama-2-7B \\
\hline
Self-Talk~\cite{selftalk} & Multi-agent & Human or Rule & / & \checkmark & LoRA & MosaicAI-7B-Chat \\
\hline
ENVISIONS~\cite{envisions} & Self-exploration & Environment & / & \checkmark & Customized & Llama2-7B/13B-Chat\\
\hline
AgentGym~\cite{agentgym}& Strong LLM\&Expert & Environment & / & \checkmark & BC & Llama-2-7B-Chat \\
\hline
FireAct~\cite{fireact} & Strong LLM & Environment & / & / & LoRA & GPT3.5, Llama-2-7B/13B, CodeLlama-7B/13B/34B-Instruct \\
\hline
NAT~\cite{nat} & Strong LLM & Environment & / & \checkmark & SFT & Llama-2-7B/13B-Chat\\
\hline
Agent Lumos~\cite{agentlumos}& Strong LLM & Human or Rule & / & / & LoRA & Llama-2-7B/13B \\
\hline
STE~\cite{STE} & Self-exploration & Model & / & \checkmark & SFT & Llama-2-7B/13B-Chat, Mistral-7B-Instruct\\
\hline
OPTIMA~\cite{optima} & Multi-agent & Human or Rule & \checkmark & / & SFT & Llama-3-8B \\
\hline
Zhou et al.~\cite{enhancing} & Strong LLM & Human or Rule & \checkmark & / & LoRA & OpenChat v3.2, Llama-2-7B, AgentLM-7B \\
\hline
AgentOhana~\cite{agentohana} & Expert & Model & / & / & QLoRA & xLAM-v0.1 \\
\hline
COEVOL~\cite{coevol} & Expert & Model & \checkmark & / & SFT & Llama-2-7B, Mistral-7B \\
\hline
AgentBank~\cite{agentbank} & Strong LLM & Environment & / & \checkmark & Instruction Tuning & Llama-2-Chat \\
\hline
AdaSwitch~\cite{adaswitch} & Self-exploration & Model  & \checkmark & \checkmark & SFT & DeepSeek-Coder-1.3B, StarCoder2-3B \\
\hline
IPR~\cite{IPR} &Expert \& Self-exploration & Environment & / & \checkmark & Instruction Tuning & Llama-2-7B \\
\hline
Re-ReST~\cite{re-rest} & Self-exploration & Environment & / & \checkmark & LoRA & Llama-2-7B/13B, Llama-3-8B, CodeLlama-13B, VPGen \\
\hline
ATM~\cite{atm} & Multi-agent & / & \checkmark & / & MITO (Customized) & Llama-2-7B \\
\hline
Aksitov et al.~\cite{rest} & Self-exploration & Model & / & / & SFT & PaLM-2-base-series \\
\hline
SWIFTSAGE~\cite{swiftsage} & Self-exploration & Environment & \checkmark & / & SFT & T5-Large \\
\hline
AGILE~\cite{agile} & Expert & / & / & / & BC & Vicuna-13B, Meerkat-7B \\
\hline
NLRL~\cite{nlrl} & Self-exploration & / & / & / & SFT & Llama-3.1-8B-Instruct \\
\hline
ETO~\cite{eto} & Expert & / & / & \checkmark & BC & Llama-2-7B-Chat \\
\hline
Retrospex~\cite{retrospex} & Expert & / & / & \checkmark & BC & Flan-T5-Large, Llama-3-8B-Instruct \\
\hline
ToRA~\cite{tora} & Strong LLM & Human or Rule & / & \checkmark & BC & Llama-2-series, CodeLlama-series \\
\hline
Sayself~\cite{sayself} & Strong LLM & Human or Rule  & / & / & SFT & Mistral-7B, Llama-3-8B \\
\hline
Star-Agents~\cite{star-agents} & Strong LLM & Model  & \checkmark & / & SFT & Pythia-1B, Llama-2-7B \\
\hline
ATLaS~\cite{atlas} & Expert  & Model  & / & / & Customized & Mistral-7B, Llama-3-8B \\
\hline
WKM~\cite{wkm} & Expert & Model  & \checkmark & / & LoRA & Mistral-7B-Instruct, Llama-3-8B-Instruct,Gemma-1.1-7B \\
\hline
Subramaniam et al.~\cite{ma-ft} & Multi-agent & Model  & \checkmark & / & SFT & Phi-3-Mini-128K-Instruct, Mistral-7B-Instruct,Llama-3-8B-Instruct\\
\hline
E$^2$CL~\cite{e2cl} & Expert\& Self-exploration  & Human or Rule & / & \checkmark & SFT & Flan-t5-small/base/large \\
\hline
STeCa~\cite{steca} & Expert & / & / & \checkmark & SFT & Llama-2-7B-Chat, Llama-3-8B-Instruct, Mistral-7B \\
\hline

\end{tabular}%

}
\end{center}
\label{ft_optimization_tab}
\end{table}

\subsubsection{\textbf{Trajectory Data Construction for Agent Fine-Tuning}} \label{trajectory data}
Previous studies~\cite{dataquality1,ftsurvey,nlrl} have shown that the quality of training data significantly impacts model performance, making trajectory construction a critical step in the fine-tuning pipeline. This process typically involves trajectory data generation, evaluation and filtering, and utilization of low-quality samples, to construct refined data that meet the requirements for effective fine-tuning.

\textbf{Data Acquisition and Generation.} High-quality trajectory construction begins with the acquisition and generation of initial data, which require not only diversity but also strong task alignment to ensure effective learning. These methods can generally be classified into four broad categories: expert-annotated data, strong LLM-generated trajectories, self-exploration environment-interaction trajectories, and multi-agent collaboration-based construction. 

\textbf{(1) \textit{Expert-annotated data}. }
Expert-annotated trajectories provide high-quality, human-curated demonstrations and are widely regarded as the gold standard for agent fine-tuning. Such data ensure strong task alignment because experts can design thoughts, observations, and actions~\cite{react} that model the intended decision-making process. Many systems~\cite{eto,IPR,agile,fireact,nat,steca} adopt ReAct-style trajectories as foundational training data, with methods like IPR~\cite{IPR}, ETO~\cite{eto}, and AGILE~\cite{agile} using CoT-enhanced demonstrations~\cite{cot} to reinforce task-specific reasoning skills. To improve compatibility with pre-trained LLM distributions, Agent-FLAN~\cite{agentflan} restructures ReAct trajectories into multi-turn dialogue formats, while AgentOhana~\cite{agentohana} further standardizes heterogeneous expert trajectories to improve consistency.

\textbf{(2) \textit{Strong LLM-generated trajectories}.} 
Strong LLM-generated trajectories rely on powerful models such as ChatGPT or GPT-4 to autonomously produce task-specific data, typically using reasoning frameworks like ReAct and CoT to simulate multi-step reasoning, decision-making, and interaction with environments. Many approaches, including AgentTuning~\cite{agenttuning} and FireAct~\cite{fireact}, employ ReAct-, CoT-, or Reflexion-style prompting~\cite{reflexion} to improve trajectory diversity and reasoning quality. Other works enrich trajectories with tool use or structured annotations, such as NAT~\cite{nat}, which incorporates calculators and APIs, and Agent Lumos~\cite{agentlumos}, which uses GPT-4/GPT-4V to produce annotated planning and grounding data. Multi-role simulation has also been explored, with systems like Zhou et al.~\cite{enhancing} using GPT-4 to coordinate problem generators, planners, and environment agents to iteratively generate more complex trajectories.

\textbf{(3) \textit{Self-exploration environment-interaction trajectories}.} Given the high cost of proprietary models like GPT-4, self-exploration with open-source LLMs has become a common strategy for generating trajectory data. These approaches allow agents to interact with environments and improve trajectories through autonomous exploration, feedback, and self-training. Early methods~\cite{swiftsage,rest} directly collect trajectories from environment interactions, such as SWIFTSAGE~\cite{swiftsage}, which samples actions based on state–history pairs to build concise datasets. More recent work incorporates feedback and reflection, with systems such as ENVISIONS~\cite{envisions} and STE~\cite{STE} enabling self-correction, tool/API simulation, and experience storage to refine generated trajectories. Other approaches use auxiliary models to guide refinement: AdaSwitch~\cite{adaswitch} employs a stronger cloud LLM to correct errors made by a local model, while Re-ReST~\cite{re-rest} iteratively improves outputs through a reflector model. Natural language–based RL frameworks like NLRL~\cite{nlrl} further unify policies, values, and feedback in text form, enabling trajectory generation from an RL perspective.


\textbf{(4) \textit{Multi-agent collaboration-based construction}.}
Multi-agent collaborative frameworks enhance trajectory diversity and robustness by incorporating role assignment and coordinated interactions that go beyond single-agent generation. Although often built on strong LLMs or self-exploration, their collaborative mechanisms merit separate attention. Most approaches~\cite{smart,coevol,atm,selftalk} assign complementary roles and let agents iteratively generate or refine trajectories through sequential or alternating interactions. For example, SMART~\cite{smart} organizes agents as intent reconstructor, knowledge retriever, and response generator to co-produce multi-turn trajectories, while COEVOL~\cite{coevol} uses advisor–editor–judge roles in a debate-style refinement loop. To improve robustness, adversarial collaboration frameworks such as ATM~\cite{atm} pair an attacker perturbing retrieved evidence with a generator trained to produce disturbance-resistant outputs.

\begin{table}[htbp]
\setlength{\abovecaptionskip}{0.1cm}
\caption{Comparison of Data Acquisition and Generation Strategies.}
\centering             

\renewcommand{\arraystretch}{1} 
\resizebox{0.9\textwidth}{!}{%
\begin{tabular}{
  >{\centering\arraybackslash}m{2.0cm} | 
  >{\centering\arraybackslash}m{6.0cm} | 
  >{\centering\arraybackslash}m{7cm}
}
\hline
\textbf{Strategy} & \textbf{Advantages} & \textbf{Limitations} \\
\hline
Expert-annotated
& Ensures high accuracy with flexible, task-specific manual design. 
& Labor-intensive, limited, and difficult to obtain in professional domains. \\
\hline
Strong LLM-generated
& Provides stable and high-performance trajectories without complex design. 
&Incurs high usage cost and may introduce potential biases and factual inconsistencies. \\
\hline
Self-exploration
& Offers low cost, easy implementation, and no need for extra assistance. 
& Easily produces low-quality trajectories with errors, requiring effective data filtering. \\
\hline
Multi-agent collaboration
& Improves accuracy and scalability through task specialization. 
& Relies on complex coordination mechanisms, which increase design complexity and cost. \\
\hline
\end{tabular}
}
\label{tab:data_acquisition_comparison}
\end{table}

In conclusion, data acquisition and generation strategies form the foundation of trajectory construction for LLM-based agents, with each approach offering distinct benefits and limitations, as summarized in Table~\ref{tab:data_acquisition_comparison}. Their suitability varies with task demands, reflecting trade-offs among quality, scalability, and complexity. Future work may benefit from hybrid strategies that combine complementary methods to better leverage their respective strengths.

\textbf{Data Evaluation and Filtering.}
After trajectory data are generated, evaluating and filtering it becomes crucial for ensuring fine-tuning quality. While expert-annotated datasets are often directly usable, other trajectory types typically require additional refinement through criteria-based assessment, external feedback, or model scoring. We group existing filtering approaches into three categories: environment-based, human or rule-based, and model-based evaluation.

\textbf{(1) \textit{Environment-based}. }
Environment-based methods rely on feedback from the environment, such as environmental rewards or task completion signals, to assess agent actions and filter trajectory data.
Most works~\cite{fireact,agenttuning,envisions,agentgym} in this category adopt a form of binary feedback, where the reward is assigned based on the success or failure of the agent’s actions relative to the desired task. AgentTuning~\cite{agenttuning}, ENVISIONS~\cite{envisions} and FireAct~\cite{fireact} use environment-based rewards to evaluate trajectory success, relying on environmental signals or correctness assessments. Similarly, NAT~\cite{nat} also employs a binary environment-based reward system, differentiating between successful and unsuccessful outcomes to encourage agents to learn from failure.


\textbf{(2) \textit{Human or Rule-based}. }
Human or rule-based evaluation assesses trajectory quality through predefined criteria, manual inspection, or task-specific rules, focusing on factors such as correctness, diversity, and consistency. Many methods~\cite{enhancing,agentlumos,rest} adopt such approaches to ensure alignment with task requirements. Typical strategies include manual filtering to verify consistency between environment feedback and agent actions~\cite{enhancing}, or perplexity-based selection to retain samples most aligned with task objectives~\cite{rest}. Other systems~\cite{selftalk,envisions,optima} employ multi-dimensional criteria—e.g., dialogue diversity, sub-goal completion, or readability to select higher-quality trajectories, with some methods ranking trajectories using reward or probability-based scoring~\cite{envisions}.


\textbf{(3) \textit{Model-based}.}
Model-based evaluation leverages LLMs (e.g., GPT-3.5/4) to automate trajectory assessment using their reasoning and scoring capabilities. Such methods typically rely on LLMs to rank or validate trajectories for relevance, correctness, and completeness. For example, \cite{rest} and AdaSwitch~\cite{adaswitch} employ LLM-based scoring to compare multiple samples and verify stepwise correctness, while STE~\cite{STE} uses GPT-4 to assess examples involving tool or API usage. Beyond input–output checks, several approaches evaluate full trajectories: AgentOhana~\cite{agentohana} applies an AgentRater to score process integrity and outcome validity, retaining only high-quality trajectories, and COEVOL~\cite{coevol} uses GPT-3.5 for comparative evaluation across criteria such as helpfulness, relevance, and accuracy.


\begin{table}[htbp]
\setlength{\abovecaptionskip}{0.1cm}
\caption{Comparison of Data Evaluation and Filtering Strategies.}
\centering             

\renewcommand{\arraystretch}{1} 
\resizebox{0.9\textwidth}{!}{%
\begin{tabular}{
  >{\centering\arraybackslash}m{2.0cm} | 
  >{\centering\arraybackslash}m{5.5cm} | 
  >{\centering\arraybackslash}m{8cm}
}
\hline
\textbf{Strategy} & \textbf{Advantages} & \textbf{Limitations} \\
\hline
Environment-based
& Easy to implement with clear and well-defined reward signals. 
& Primarily focuses on final binary outcomes, neglecting step feedback and potentially accumulating errors. \\
\hline
Human or Rule-based
& High reliability, strong interpretability, and flexible customization.
& Relies on predefined metrics or human oversight, and requires complex rule design. \\
\hline

Model-based
& Fully automated process, reducing human effort and manual design.
& Depends on model reliability and is affected by inherent biases within the model. \\
\hline
\end{tabular}
}
\label{tab:data_fiter}
\end{table}

In summary, different evaluation and filtering strategies vary in their advantages and limitations, as summarized in Table~\ref{tab:data_fiter}. 
These methods offer distinct trade-offs in scalability, accuracy, and implementation complexity, requiring careful selection and design based on agent task-specific demands. Integrating multiple approaches into a hybrid strategy can further enhance data quality and robustness, ensuring a more reliable evaluation framework tailored to different task needs.


\textbf{Low-quality Data Utilization}.
Because filtering often yields limited high-quality data, leveraging low-quality or failed trajectories has become an effective strategy for dataset augmentation. Early approaches relied primarily on expert-crafted or successful trajectories, overlooking the informative value of failures. Subsequent methods address this in two ways.
Several works~\cite{agentgym,IPR,epo} construct comparative datasets by pairing successful and failed trajectories, enabling agents to learn from both correct and incorrect behaviors. Others focus on refining low-quality samples: AgentBank~\cite{agentbank} regenerates failed interactions, while Re-ReST~\cite{re-rest} and AdaSwitch~\cite{adaswitch} iteratively correct errors using environment or model feedback.
In contrast, some approaches use error samples directly to teach failure recognition. Agent-FLAN~\cite{agentflan} integrates frequent negative examples to reduce hallucinations, and NAT~\cite{nat} generates deliberately incorrect responses to strengthen the model’s ability to distinguish valid from invalid outputs.
 

\subsubsection{\textbf{Trajectory-based Fine-Tuning for Agent}}
Fine-tuning is a critical step in the optimization process for LLM-based agents, enabling open LLMs to adapt to specific agent tasks or data distributions. Most techniques are based on standard LLM fine-tuning methods, but are specifically applied to agent-related data (constructed in §\ref{trajectory data}) to enhance the decision-making and task completion. 

\textbf{Mixing of General Instructions and Agent Trajectories}. Some studies~\cite{enhancing,agentflan} have shown that fine-tuning solely on agent-specific trajectories may weaken general language and reasoning abilities of LLM-based agents, so it is common to mix general instruction datasets and task-specific trajectories during fine-tuning to preserve foundational skills while optimizing for agent-specific tasks.
For example, AgentTuning~\cite{agenttuning} combines AgentInstruct datasets with open-source instructions to fine-tune LLaMA-2 models, while AgentBank~\cite{agentbank} integrates agent trajectories, general instruction data, and code datasets to optimize LLaMA-2-Chat models for diverse tasks.

\textbf{The Usage of Fine-Tuning Techniques}.
Based on studies on LLM-based agents, we categorized fine-tuning methods into three types: standard SFT, PEFT (e.g., LoRA), and customized strategies tailored for specialized tasks, which will be discussed in detail.

\textbf{(1)\textit{Standard SFT}.}
The goal of SFT in LLM-based agents is to align pre-trained models with specific task requirements by minimizing the discrepancy between predicted and target outputs. Here, standard SFT refers to full-parameter fine-tuning on high-quality training data, including instruction tuning. Since Behavior Cloning (BC) in imitation learning follows the same paradigm, we also include it in this category.
Most works adopt standard SFT to fine-tune LLM-based agents using high-quality datasets. AgentTuning~\cite{agenttuning}, Agent-FLAN~\cite{agentflan}, and AgentBank~\cite{agentbank} fine-tune Llama-based models with instruction data to align them with task requirements. Similarly, NAT~\cite{nat}, STE~\cite{STE}, and COEVOL~\cite{coevol} employ SFT on trajectory datasets capturing both successful and failed task cases. In addition, AGILE~\cite{agile}, Retrospex~\cite{retrospex}, AgentGym~\cite{agentgym}, ETO~\cite{eto}, and ToRA~\cite{tora} adopt BC-style SFT to equip models with foundational agent capabilities.
Overall, the simplicity and effectiveness of standard SFT make it a core method for aligning LLMs with task-specific behaviors.

\textbf{(2)\textit{Parameter-efficient fine-tuning}.}
PEFT methods, such as LoRA and QLoRA, update only a small number of parameters while keeping most of the LLM frozen, significantly reducing computational and memory costs and enabling efficient fine-tuning of large models.
Many works, including SMART~\cite{smart}, FireAct~\cite{fireact}, Re-ReST~\cite{re-rest}, and Agent Lumos~\cite{agentlumos}, apply LoRA to fine-tune models from the Llama series, OpenChat, and CodeLlama across various configurations. AgentOhana~\cite{agentohana} further uses QLoRA to optimize xLAM-v0.1 models, demonstrating effective fine-tuning under resource constraints.
Overall, PEFT provides a practical solution for optimizing LLM-based agents, balancing between performance and efficiency in limited-resource settings.

\textbf{(3)\textit{Customized fine-tuning}.}
Customized fine-tuning methods are tailored to specific tasks or ideas, incorporating designed strategies or objective functions,  such as introducing additional regularization constraints, to better align agents with task requirements.
ATM~\cite{atm} designs Multi-agent Iterative Tuning Optimization (MITO) loss, which combines standard SFT with KL regularization to balance task-specific optimization and generalization. Leveraging contrastive learning and RL-free optimization, ENVISIONS~\cite{envisions} achieves efficient iterative refinement of Llama-2 through cyclic fine-tuning on dynamically updated trajectory pairs.
These methods highlight the flexibility of adapting LLMs to agent-specific tasks by customized objectives and iterative optimization.

\subsubsection{\textbf{Summary}} Conventional fine-tuning-based optimization has effectively enhanced LLM-based agents by high-quality trajectory data, different training strategies, and efficient parameter updates, ensuring strong task alignment, controlled optimization, and adaptability to specific objectives.
However, such methods also face inherent limitations due to their dependence on curated data, vulnerability to overfitting, and limited adaptability to dynamic environments. Additionally, without interactive feedback mechanisms, they may accumulate errors and struggle with generalization, because fine-tuning typically aligns models with static objectives and restricts real-time behavioral refinement. With the growing complexity and dynamics of agent tasks, exploring more adaptive optimization strategies becomes essential to further enhance their decision-making capabilities.

\subsection{Reinforcement Learning-based Optimization}\label{sec3.2}
\begin{figure}[htbp]
\centerline{\includegraphics[width=0.9\linewidth]{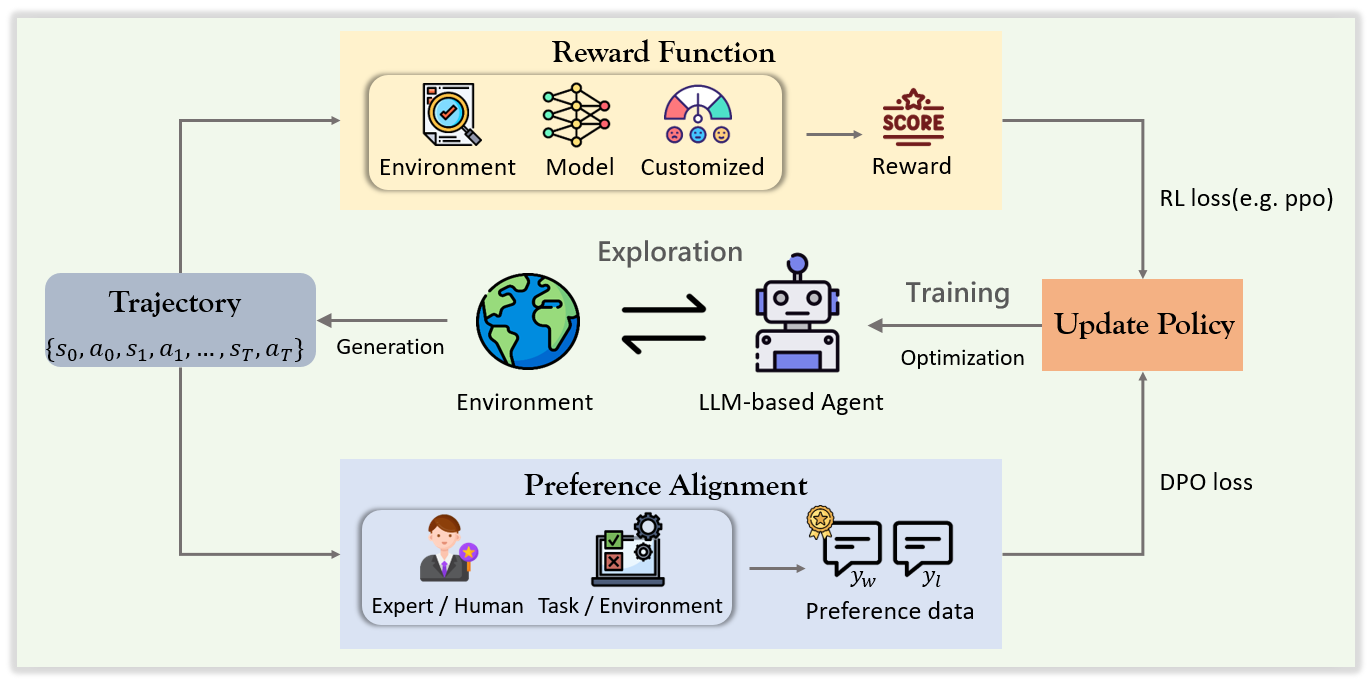}}
\setlength{\abovecaptionskip}{0.1cm}
\caption{Workflow of Reinforcement Learning-based Optimization for LLM-based Agents.}
\label{fig:rltuning}
\end{figure}

RL-based optimization has emerged as a promising approach, enabling LLM-based agents to learn directly from interactions with environment or human feedback and dynamically refine their actions through rewards. Unlike static fine-tuning, RL encourages exploratory learning, allowing agents to discover novel policies and adapt to unseen tasks. By aligning model outputs with explicit rewards or preferences, RL enhances both task performance and robustness in complex, evolving environments. We categorize RL-based optimization methods into two primary approaches: \textbf{Reward-function-based optimization} and \textbf{Preference-alignment-based optimization}. 
The overall workflow is illustrated in Figure~\ref{fig:rltuning}.

\vspace{-1em}

\subsubsection{\textbf{Reward Function-based Optimization}}
Reward function-based optimization methods leverage explicit reward signals to refine the behavior of LLM-based agents, facilitating adaptation to complex and dynamic tasks. Drawing from traditional RL paradigms, these methods employ algorithms like PPO to iteratively optimize agents' policies by adjusting LLMs' parameters. By treating LLM as an agent, these approaches utilize diverse reward sources, including environmental feedback, model-generated signals, and customized reward functions, as summarized in Table~\ref{tab_rl_methods}.
\begin{table}[htbp]
\setlength{\abovecaptionskip}{0.1cm}
\caption{Summary of Reward Function-based Optimization Methods for LLM-based Agents. Note: MA- Multi-Agent Framework.}
\begin{center}
\resizebox{0.9\textwidth}{!}{%
\begin{tabular}{c|c|c|c|c}
\hline
\textbf{Method} & \textbf{Reward Source} & \textbf{RL Algorithm} & \textbf{MA} & \textbf{Base Model} \\
\hline
CMAT~\cite{cmat} & Environment-based & Actor-Critic & \checkmark & Tinyagent-1.8B/7B \\
\hline
StepAgent~\cite{stepagent} & Model-based & DPO+PPO & / & Mistral-7B, Llama3-8B \\
\hline
WebRL~\cite{webrl} & Model-based & Actor-Critic & \checkmark & GLM-4-9B, Llama-3.1-8b/70B \\
\hline
CORY~\cite{cory} & Customized, Model-based & PPO & \checkmark & GPT 2-Large, Llama-2-7B-Chat \\
\hline
SaySelf~\cite{sayself} & Customized & PPO & \checkmark & Mistral-7B, Llama-3-8B \\
\hline

AgentGym~\cite{agentgym} & Environment-based & AgentEvol & / & Llama-2-7B-Chat\\
\hline
GELI~\cite{geli} &Model-based & PPO & / & Llama-2-7B\\
\hline
AGILE~\cite{agile} & Customized & PPO & / & Vicuna-13B, Meerkat-7B \\
\hline
RISE~\cite{rise} & Environment-based & RWR & / & Llama-2-7B-chat, Mistral-7B-Instruct \\
\hline
KALM~\cite{kalm} & Environment-based & CQL & / & Llama-2-7B-Chat-hf \\
\hline
RL4VLM~\cite{rl4vlm} & Environment-based & PPO & / & Llava-v1.6-mistral-7B \\
\hline
POAD~\cite{POAD} & Environment-based & POAD & / & Llama-3.1-8B-Instruct \\
\hline
STeCa~\cite{steca} & Customized, Model-based & PG & / & Llama-2-7B-Chat, Llama-3-8B-Instruct, Mistral-7B \\
\hline
DeepResearcher~\cite{zheng2025deepresearcher}& Customized & GRPO & / & Qwen2.5-7B-Instruct\\
\hline
\end{tabular}
}
\end{center}
\label{tab_rl_methods}
\end{table}

\textbf{Environment-based rewards}. One of the most common approaches leverages environmental feedback as the primary reward source. CMAT~\cite{cmat} uses an actor–critic framework in multi-agent collaboration, where agents interact with environments and receive rewards based on task outcomes. Retrospex~\cite{retrospex} applies offline implicit Q-Learning to stabilize policy updates by minimizing temporal-difference (TD) error on fixed trajectories. AgentGym~\cite{agentgym} optimizes itself using the AgentEvol algorithm guided by environment rewards.
Environment-based rewards are straightforward to obtain, but are often discrete and outcome-focused, which can limit their ability to provide detailed guidance during intermediate steps of complex tasks.

\textbf{Model-based rewards}. Model-based rewards derive signals from models trained to evaluate agent behaviors, providing implicit or explicit feedback to guide policy optimization. These methods are particularly useful when direct environmental rewards are sparse or unavailable. StepAgent~\cite{stepagent} employs an inverse RL framework in which a discriminator predicts how well trajectories align with expert behaviors and provides fine-grained step-wise rewards for PPO optimization. WebRL~\cite{webrl} trains a self-supervised Optimal Reward Model to evaluate trajectory quality; using an actor–critic framework, it updates policy weights with ORM-generated rewards and applies KL constraints for policy stability.
In summary, model-based rewards offer fine-grained, step-level feedback that enhances generalization beyond final outcome rewards. However, their effectiveness depends on the robustness of the reward models, which require careful design and optimization.

\textbf{Customized reward functions}. Customized reward functions adapt reinforcement learning frameworks to the specific objectives of agent tasks, extending optimization signals beyond binary task success to factors such as stability, efficiency, and reliability. GELI~\cite{geli} integrates global rewards with local multimodal feedback to guide long-horizon conversational behavior using PPO. AGILE~\cite{agile} designs a reward balancing task completion and expert-query costs, encouraging strategic assistance-seeking. CORY~\cite{cory} employs shared rewards and KL constraints in a multi-agent setting to improve policy quality, while SaySelf~\cite{sayself} calibrates model confidence by rewarding accurate predictions and penalizing overconfident errors.
Recent advances extend these ideas to more complex reasoning. \textcolor{black}{DeepResearcher~\cite{zheng2025deepresearcher} introduces an end-to-end RL framework in real web environments, enabling LLM agents to iteratively reason, search, and synthesize information for open-domain questions.} Collectively, these works illustrate the flexibility of RL-based optimization through task-specific reward designs tailored to diverse agent capabilities.


\subsubsection{\textbf{Preference Alignment-based Optimization}}

Preference alignment-based optimization offers an alternative to traditional RL by aligning model behavior with human or other preferences without relying on explicit reward signals. Algorithms such as DPO can be viewed as RL variants tailored for offline datasets, where preferences replace reward signals to guide policy refinement. This offline formulation enables effective tuning of large-scale LLM-based agents without online rollout.
The optimization process typically consists of two stages: \textbf{Preference Data Construction} and \textbf{Policy Optimization}, as follows.

\begin{itemize}
\item \textbf{Preference Data Construction.}
The first step is constructing preference data in the form of pairwise comparisons, represented as tuples $(x, y_w, y_l)$, where $x$ is the input instruction, $y_w$ denotes the preferred (win) response, and $y_l$ is the less preferred (lose) response (Some studies alternatively use $a^+$ and $a^-$). This structured format explicitly ranks outputs, forming a clear basis for preference-based optimization. Methods such as Monte Carlo Tree Search (MCTS) are often employed to generate diverse trajectories, which are then evaluated using task success rates, environment feedback, or other metrics to determine $y_w$ and $y_l$.

\item \textbf{Policy Optimization via DPO.}
Following the construction of preference data, the agent's policy is optimized to align with these preferences. DPO is a widely adopted approach that directly defines a preference loss based on policy, optimizing LLM-based agents to prioritize desired outputs and adhere to task-specific requirements.
The optimization objective of DPO is formalized as follows:
\begin{equation}
\mathcal{L}_{DPO}(\pi_\theta; \pi_{ref}) = 
- \mathbb{E}_{(x, y_w, y_l) \sim \mathcal{D}} 
\left[ 
\log \sigma \left( 
\beta \log \frac{\pi_\theta(y_w \mid x)}{\pi_{ref}(y_w \mid x)} 
- 
\beta \log \frac{\pi_\theta(y_l \mid x)}{\pi_{ref}(y_l \mid x)} 
\right) 
\right]
\label{eq_dpo}
\end{equation}
where $D$ is the preference dataset, $\pi_\theta(y\mid x)$ denotes the probability assigned by the current policy to output $y$ given input $x$, and $\pi_{ref}(y \mid x)$ is output probability under a frozen reference model. The regularization term $\beta$ ensures the updates remain consistent with the original policy and avoids significant deviations from the reference model, while $\sigma$ is the sigmoid function ensuring stable and efficient optimization.
\end{itemize}

We summarize representative works under this paradigm, categorized by preference criteria, including those based on human and task feedback, as shown in Table \ref{tab_dpo_methods}.

\begin{table}[htbp]
\setlength{\abovecaptionskip}{0.1cm}
\caption{Summary of Preference Alignment-based Optimization Methods for LLM-based Agents. Note: MA- Multi-Agent Framework.}
\begin{center}
\resizebox{0.9\textwidth}{!}{%
\begin{tabular}{c|c|c|c|c}
\hline
\textbf{Method} & \textbf{Preference Criteria} & \textbf{Algorithm} & \textbf{MA} & \textbf{Base Model} \\
\hline
StepAgent~\cite{stepagent} & Expert  & DPO+PPO & / & Mistral-7B, Llama-3-8B \\
\hline
Agent Q~\cite{agentq} & Environment & DPO &/ & Llama-3-70B, xLAM-v0.1-r \\
\hline
ETO~\cite{eto} & Environment & DPO &/ & Llama-2-7B-Chat \\
\hline
OPTIMA~\cite{optima} & Task, Human  &  DPO & \checkmark &Llama-3-8B\\
\hline
IPR~\cite{IPR} & Expert,Environment & DPO & / & Llama-2-7B \\
\hline
Re-ReST~\cite{re-rest} & Environment & DPO &  \checkmark  & Llama-2-7/13B, Llama-3-8B, Codellama-13B, VPGen\\
\hline
DMPO~\cite{dmpo} & Expert & DMPO & / & Llama-2-7B-Chat, Mistral-7B-Instruct-v0.2 \\
\hline
EPO~\cite{epo} &Task, Environment & DPO &\checkmark& Llama-2-7B\\
\hline
ATM~\cite{atm} & Human & DPO & \checkmark & Mistral-7B-Chat \\
\hline
AMOR~\cite{amor} & Expert, Human &KTO &\checkmark & Llama-2-7B/13B-Chat \\
\hline
DITS~\cite{DITS} & Environment, Human & DPO &\checkmark & Llama-3-8B-Instruct \\
\hline
\end{tabular}
}
\end{center}
\label{tab_dpo_methods}
\end{table}

\textbf{Optimization Based on Expert or Human Preferences.}
Preference data in this category rely on expert trajectories or human-defined rules, constructing positive samples $y_w$ from ideal or successful behaviors and negative samples $y_l$ from suboptimal trajectories.
DMPO~\cite{dmpo} refines RL objectives by introducing state–action occupancy constraints from expert trajectories, guiding agents to mimic expert distributions; with length normalization, DMPO directly maximizes the likelihood of preferred trajectories to reduce multi-turn errors. IPR~\cite{IPR} builds step-level preference data by comparing agent actions with expert demonstrations and optimizes both outcome-DPO and step-DPO losses for fine-grained alignment.
AMOR~\cite{amor} adopts a two-stage approach, combining SFT pre-training with Kahneman–Tversky Optimization (KTO). In the adaptive stage, human-annotated binary feedback is used to optimize module-specific parameters.
Although expert and human-based preferences ensure high-quality data and accurate optimization, they are limited in scalability and coverage, as expert trajectories may not sufficiently represent diverse or unexplored scenarios in complex environments.

\textbf{Optimization Based on Task or Environment Preferences.}
This category constructs preference datasets using task-specific metrics or environment feedback, leveraging rewards, success rates, or other performance indicators to form preference pairs for DPO optimization. Incorporating environment-specific signals improves adaptability to dynamic and task-oriented contexts.
Methods such as ETO~\cite{eto} and Re-ReST~\cite{re-rest} directly use task success or environment feedback to build preference data. EPO~\cite{epo} ranks outputs via a reward model and combines DPO with token-level alignment loss for stable training and precise task alignment.
Structured methods include Agent Q~\cite{agentq}, which uses MCTS to explore action trajectories and generate preferences from process feedback and success rates. OPTIMA~\cite{optima} ranks MCTS node pairs by common ancestry and reward differences, selecting top-ranked pairs for DPO optimization. DITS~\cite{DITS} adopts a similar strategy and introduces a data influence metric to jointly score node pairs.


\subsubsection{\textbf{Summary}} RL-based optimization methods have shown significant potential in enhancing LLM-based agents. By incorporating reward signals or preference alignment, these methods enable agents to adapt to complex and dynamic tasks. The choice of method often depends on task complexity and data availability: reward-function-based techniques excel in challenging scenarios but require substantial data and computation, whereas preference-alignment approaches simplify training and better match expected outputs but depend heavily on the quality and coverage of preference data, limiting adaptability in highly dynamic settings. Designing effective reward functions or preference pairs also requires careful consideration to capture task-relevant factors and avoid overly narrow criteria. These challenges underscore the need for progress in algorithmic efficiency and reward signal design.
\textcolor{black}{Moreover, current RL approaches remain limited in supporting long-horizon, multi-turn interactions, highlighting the need for more general and scalable training pipelines. Recent frameworks such as AgentGym-RL~\cite{agentgym-rl} reflect this direction by providing unified environments and multi-turn RL interfaces that enable more reliable long-horizon decision-making for LLM-based agents.}

\subsection{Hybrid Fine-Tuning Optimization}\label{sec3.3}
While traditional SFT provides a stable method for initializing LLM-based agents, it often struggles with dynamic tasks requiring complex decision-making. On the other hand, RL excels in exploring complex scenarios but typically demands extensive data and computational resources. To overcome these limitations, hybrid fine-tuning optimization combines the strengths of both SFT and RL, creating a more flexible and effective framework.
This section introduces representative works that exemplify these hybrid approaches, as summarized in Table~\ref{tab_hybrid_methods}. We categorize them into two groups: sequential hybrid fine-tuning methods and non-sequential hybrid fine-tuning methods.
\begin{table}[htbp]
\setlength{\abovecaptionskip}{0.1cm}
\caption{Summary of Hybrid Fine-tuning Optimization Methods for LLM-based Agents. Note: MA- Multi-Agent Framework.}
\begin{center}
\resizebox{\textwidth}{!}{%
\begin{tabular}{c|c|c|c|c|c}
\hline
\textbf{Method} & \textbf{Fine-tune Algorithm} & \textbf{RL algorithm} & \textbf{Hybrid Strategy} & \textbf{MA} & \textbf{Base Model} \\
\hline
ETO~\cite{eto} & BC & DPO & Sequential & / & Llama-2-7B-Chat \\
\hline
AGILE~\cite{agile} & BC & PPO & Sequential & / & Vicuna-13B, Meerkat-7B \\
\hline
AMOR~\cite{amor} & SFT  & KTO  & Sequential & \checkmark & Llama-2-7B/13B-Chat \\
\hline
STeCa~\cite{steca} & SFT  & PG  & Sequential & / & Llama-2-7B-Chat, Llama-3-8B-Instruct, Mistral-7B \\
\hline
SaySelf~\cite{sayself} & SFT  & PPO & Sequential & \checkmark & Mistral-7B, Llama-3-8B \\
\hline
RL4VLM~\cite{rl4vlm} & SFT  & PPO & Sequential & / & Llava-v1.6-mistral-7B \\
\hline
OPTIMA~\cite{optima} & SFT &  DPO & Non-sequential & \checkmark & Llama-3-8B \\
\hline
IPR~\cite{IPR} & SFT  & DPO & Non-sequential & / & Llama-2-7B \\
\hline
\end{tabular}
}
\end{center}
\label{tab_hybrid_methods}
\end{table}

\textbf{Sequential hybrid fine-tuning.}
Most works in hybrid fine-tuning follow a sequential approach, starting with a warm-up stage where behavior cloning-based SFT is applied to equip LLMs with foundational capabilities using high-quality datasets, such as expert trajectories. In the subsequent RL stage, algorithms like PPO or DPO refine the agent’s policy for task-specific objectives or dynamic environments. This paradigm, similar to Reinforcement Fine-Tuning (RFT) introduced by OpenAI, has gained widespread popularity for combining SFT and RL to enhance LLMs' capabilities.
For example, ReFT~\cite{reft}, ETO~\cite{eto}, Re-ReST~\cite{re-rest}, AGILE~\cite{agile}, and AMOR~\cite{amor} all utilize SFT to train models for the warm-up phase. Following initialization, these methods apply various RL strategies. ETO and Re-ReST use DPO for preference alignment, AGILE employs PPO for decision-making, and AMOR applies KTO with binary feedback to align outputs with human preferences.

\textbf{Non-sequential hybrid fine-tuning.}
Other studies adopt non-sequential strategies to refine agent's policy, including alternating and dynamically adjusted optimization between SFT and RL phases. OPTIMA~\cite{optima} alternates SFT and DPO in the iterative training process, named as iSFT-DPO, enabling LLM-based agent to learn from optimal trajectories through SFT while refining its understanding using DPO based on comparative preferences. Similarly, IPR~\cite{IPR} incorporates step-level rewards, starting with ReAct-style expert trajectories for SFT and iteratively refining the policy through Outcome-DPO and Step-DPO losses. 
       

\textbf{Summary.} Hybrid fine-tuning strategies combine the strengths of SFT and RL, allowing LLM-based agents to balance structured guidance with adaptive optimization. Although these methods offer greater flexibility for complex and dynamic tasks, they also face challenges such as increased computational costs and reliance on high-quality preference or reward data.
Future research should explore more efficient hybrid designs to enhance the performance and versatility of LLM-based agents, making this an essential direction to advance agent optimization.
\section{Parameter-free Optimization of LLM-based Agents} \label{sec4}

Besides parameter-driven optimization methods critical for evolving LLM-based agents, parameter-free optimization offers a promising alternative by optimizing the agent's behavior without modifying LLM parameters. It improves performance through prompt engineering techniques or incorporating various information, such as feedback, tools and external knowledge, making it especially efficient in resource-constrained environments.
Unlike other surveys~\cite{memsurvey,agentplan,riseagent} that categorize methods according to agent structure or architecture, we categorize the methods based on the strategies and techniques employed to optimize agent behaviors, as summarized in Table~\ref{tab_parameter_free_optimization}.

\begin{table}[htbp]

\setlength{\abovecaptionskip}{0.1cm}
  \begin{center}
\caption{Comparison of Parameter-Free Optimization Methods.}
\label{tab_parameter_free_optimization}
\centering
\resizebox{\textwidth}{!}{%
\begin{tabular}{m{2.2cm}|m{3cm}|m{5.2cm}|>{\centering\arraybackslash}m{9cm}} 
\hline
\multicolumn{2}{c|}{\Large\textbf{Optimization Strategy}} & \centering \Large\textbf{Description} & \Large\textbf{Example Works }\\
\hline
\multicolumn{2}{c|}{\centering Experience-based} 
& \centering Learning from past interactions and historical experience 
& Optimus-1~\cite{optimus-1}, Experiential Co-Learning~\cite{experiential}, Expel~\cite{expel}, AutoManual~\cite{automanual}, AutoGuide~\cite{autoguide}, Agent Hospital~\cite{agenthospital} \\ 
\hline

\multirow{5}{*}{\centering Feedback-based} 
& \centering Self-Reflection 
& \centering Self-evaluating and correcting to refine current behavior 
& Reflexion~\cite{reflexion}, QueryAgent~\cite{queryagent}, SAGE~\cite{sage}, Agent-pro~\cite{agentpro}, Symbolic Learning~\cite{symbolic}, NLRL~\cite{nlrl} \\

& \centering External Feedback 
& \centering Modifying behaviors via external signals and critiques
& Retroformer~\cite{retroformer}, COPPER~\cite{COPPER}, InteRecAgent~\cite{InteRecAgent}, CoE~\cite{coe}, MPO~\cite{mpo}, WKM~\cite{wkm} \\

& \centering Meta-Prompt 
& \centering Iteratively refining global instructions/prompts
& MetaReflection~\cite{metareflection}, Retroformer~\cite{retroformer}, OPRO~\cite{llmoptimizer}, Symbolic Learning~\cite{symbolic}, FinCon~\cite{fincon}, Mass~\cite{mass}, MPO~\cite{mpo} \\
\hline
\multicolumn{2}{c|}{Tool-based} &\centering Using external tools and optimizing selection strategies& TPTU~\cite{tptu}, AVATAR~\cite{avatar}, Lyra~\cite{lyra}, Middleware~\cite{middleware}, AgentOptimizer~\cite{offline}, VideoAgent~\cite{videoagent}, AutoAct~\cite{autoact} \\ 
\hline
\multicolumn{2}{c|}{Retrieval-based} &\centering Retrieving and incorporating external knowledge& AutoRAG~\cite{autorag}, Self-RAG~\cite{self-rag}, RaDA~\cite{rada}, MALADE~\cite{malade}, EMG-RAG~\cite{emg-rag}, WebThinker~\cite{li2025webthinker}, FlowSearch~\cite{hu2025flowsearch}\\ 
\hline
\multicolumn{2}{c|}{Multi-Agent collaboration} & \centering Multiple role-specific agents communicate and collaborate & MetaGPT~\cite{metagpt}, ChatDev~\cite{chatdev}, DyLAN~\cite{dylan}, MacNet~\cite{macnet}, Agentverse~\cite{agentverse}, CAPO~\cite{capo}, SMoA~\cite{smoa}, MAD~\cite{MAD}, AutoGen~\cite{autogen}, Yuksel et al.~\cite{multiai} \\ 
\hline
\end{tabular}%
}
 \end{center}
\end{table}
\vspace{-1em} 

\subsection{Experience-based Optimization}
Experience-based optimization leverages historical data, trajectories, or accumulated knowledge to improve LLM-based agents by deriving insights from past successes and failures, often organized through memory modules. This approach refines strategies, enhances long-term decision-making, and enables agents to generalize across tasks and domains.

Optimus-1~\cite{optimus-1} utilizes a multimodal memory module to convert exploration trajectories into hierarchical knowledge graphs, which assists in task planning and prompt generation for the agent. Agent Hospital~\cite{agenthospital} integrates a medical record library and an experience repository to refine guidelines based on successful and failed cases.
To extract actionable insights, methods like ExpeL~\cite{expel}, AutoManual~\cite{automanual}, and AutoGuide~\cite{autoguide} autonomously collect, organize, and retrieve experiential knowledge to guide future actions during inference. AutoMLGen~\cite{du2025automlgen} propose MCGS to obtain more experiential solutions by integrating information from different branches.
In addition, Experiential Co-Learning~\cite{experiential} introduces a framework where instructor and assistant agents collaboratively gather shortcut-oriented experiences from their historical trajectories. These experiences are reused to inform future task execution, enabling agents to iteratively improve through shared and accumulated knowledge.

\subsection{Feedback-based Optimization}
Feedback-based optimization enhances LLM-based agents by incorporating various forms of feedback for self-reflection, correction, and iterative improvement. \textcolor{black}{These approaches operate without modifying model parameters and thus complement fine-tuning and RL by providing lightweight, inference-time mechanisms for adaptive behavior adjustment.} We categorize work into three types based on feedback strategies and their role in triggering iterative improvements: feedback-based self-reflection optimization, external feedback-based optimization, and meta-prompt optimization.

\subsubsection{\textbf{Feedback-based Self-Reflection Optimization}}
This category focuses on agent’s reflection ability using feedback from the environment or self-evaluation. The agent uses feedback to identify areas for improvement and adjust its behavior through self-correction and evolution, enhancing its adaptability and performance in dynamic environments.

Reflexion~\cite{reflexion} and QueryAgent~\cite{queryagent} leverage task outcomes or heuristic evaluations, converting them into textual feedback to guide self-reflective adjustments. In SAGE~\cite{sage}, the checker agent provides iterative feedback on the current solution, while the assistant agent generates self-reflection to support continual refinement. Agent-pro~\cite{agentpro} reflects on past trajectories and beliefs to correct suboptimal strategies.
Additionally, Symbolic Learning~\cite{symbolic} and NLRL~\cite{nlrl} use natural language to mimic optimization processes such as gradient updates or reward shaping, treating these evaluations as feedback that incrementally improves policy execution. These approaches illustrate how inference-time reflection can emulate aspects of parameter-level optimization while remaining more flexible and interpretable.

\subsubsection{\textbf{External Feedback-based Optimization}}
In this part, evaluation signals like reflections and corrections from external models or frameworks are applied to refine the behavior of the actor agent. Inspired by Actor-Critic algorithm, these methods enhance the robustness and adaptability of LLM-based agents and are also commonly used in multi-agent collaborative frameworks.

Retroformer~\cite{retroformer} and WKM~\cite{wkm} train a retrospection model and a world knowledge model to analyze failures and support global planning respectively, enabling agent to iteratively adjust actions. Similarly, COPPER~\cite{COPPER} employs a shared reflector module to generate counterfactual feedback, refine prompts, and store improvements in a memory module. InteRecAgent~\cite{InteRecAgent} applies a critic LLM checking the actor agent’s behavior to prevent instruction violations or incorrect tool usage.
In CoE~\cite{coe}, a multi-agent framework is proposed to address complex operations research, where a conductor orchestrates task assignment among specialized agents, and an evaluator provides feedback to iteratively refine the outputs through backward optimization.

\subsubsection{\textbf{Meta-Prompt Optimization}}
Meta-prompt optimization refines global instructions or meta-prompts to enhance the generalization capabilities of LLM-based agents. By iteratively adjusting prompts based on feedback, agents can better adapt to diverse and dynamic tasks, optimizing their behaviors from a broader and more general perspective.

MetaReflection~\cite{metareflection} and OPRO~\cite{llmoptimizer} extract information from failed trials or analyze task accuracy to create optimized prompts, iteratively refining instructions to align agent behavior with task objectives. Mass~\cite{mass} extends prompt optimization to a hierarchical framework, combining local prompt adjustments, workflow restructuring, and global refinements to support more stable multi-step reasoning. \textcolor{black}{MPO~\cite{mpo} fine-tunes a meta-planner through SFT and DPO to guide high-level planning via meta-prompts, illustrating how prompt-level and parameter-level optimization can work together.
Other approaches such as Symbolic Learning~\cite{symbolic} and FinCon~\cite{fincon} simulate optimization processes like gradient descent or loss shaping in natural language, enabling agents to iteratively adjust strategies without modifying underlying model parameters. These methods show how meta-prompt optimization can introduce structured behavioral priors that reinforce or extend the capabilities obtained through parameter-based training.}

\subsection{Tool-based Optimization}
One of the key distinctions between LLMs and agents is their ability to utilize external tools. These tools, such as search engines, code interpreters and domain-specific modules, enable agents to perform tasks more autonomously. Through effective selection and utilization of tools, agents can significantly expand their problem-solving scope and capabilities.

TPTU~\cite{tptu} enhances task planning and tool usage by optimizing task decomposition and tool invocation. AVATAR~\cite{avatar} employs a comparator to analyze performance differences between sample pairs, attributing discrepancies to tool usage issues and providing actionable improvements. Middleware~\cite{middleware} introduces error feedback mechanisms to align tool inputs and outputs across steps, mitigating execution errors and improving system robustness.
Additionally, AgentOptimizer~\cite{offline} optimizes tool usage by treating functions as learnable weights, then iteratively refines function sets based on execution outcomes without modifying core LLM parameters.
Other methods prioritize efficient tool integration in dynamic environments. For instance, VideoAgent~\cite{videoagent} employs a minimal yet sufficient tool set, incorporating specialized visual models and object memory queries. AutoAct~\cite{autoact} automates task trajectory generation using a pre-assembled library of tools, such as search engines and code interpreters.

\subsection{Retrieval-based Optimization}
Retrieval-based approaches enhance LLM-based agents by integrating external evidence into the reasoning process, enabling models to supplement fixed pre-trained knowledge with dynamically acquired information from heterogeneous sources. By grounding decisions in retrieved context, these methods improve factual accuracy and adaptability in evolving environments. \textcolor{black}{Beyond early retrieval-augmented techniques such as RAG, recent systems like Deep Research (DR)~\cite{xu2025drsurvey} further advance this paradigm through multi-step and iterative retrieval and report generation.}


Early approaches primarily optimize retrieval configurations or integrate retrieval with generative processes. For example, AutoRAG~\cite{autorag} explores combinations of retrieval, re-ranking, and expansion strategies to identify effective pipelines, while Self-RAG~\cite{self-rag} couples retrieval with self-reflection to iteratively refine generated content. Other systems leverage retrieval to guide task planning and execution: RaDA~\cite{rada} incorporate dynamic retrieval and past experiences to support decomposition and action generation, and PaperQA~\cite{paperqa} extends retrieval-augmented techniques to scientific domains by extracting fine-grained evidence from full-text articles. Multi-agent frameworks such as MALADE~\cite{malade} further apply retrieval to enhance coordinated decision-making, while EMG-RAG~\cite{emg-rag} learns to select memory via reinforcement learning to provide more informative contextual prompts.
\textcolor{black}{Building on these foundations, recent DR agents~\cite{xu2025drsurvey, shi2025dualresearch} significantly extend the retrieval paradigm through multi-step search gathering and structured reasoning. WebThinker~\cite{li2025webthinker} introduces a deep web exploration workflow that autonomously performs reasoning–search–synthesis cycles to generate research reports, strengthened by RL to improve tool calling capability. FlowSearch~\cite{hu2025flowsearch} develops a dynamic, multi-agent knowledge-flow framework that enables parallel exploration, recursive evidence integration, and adaptive workflow optimization for complex research-oriented tasks.}

\subsection{Multi-Agent Collaborative Optimization}\label{ma optimization}
Multi-agent collaborative optimization enhances LLM-based systems by distributing roles, coordinating information flow, and enabling agents to jointly plan and execute tasks. Through effective collaboration, these frameworks enhance collective decision-making in LLM-based multi-agent systems (MAS), achieving performance beyond the capabilities of individual agents.

Early systems such as MetaGPT~\cite{metagpt} and ChatDev~\cite{chatdev} formalize human-inspired workflows for software engineering by assigning distinct roles for planning, coding, debugging, and documentation. Recent programming-oriented agents extend this paradigm further. \textcolor{black}{For example, Claude Code provides a general-purpose multi-agent coding framework in which multiple specialized subagents coordinate long-horizon tasks through structured planning, execution, and memory-assisted refinement, enabling AI to operate as an engineering team.}
To improve coordination and organizational efficiency, DyLAN~\cite{dylan} and MacNet~\cite{macnet} dynamically construct agent networks using importance scoring and early-stopping signals to streamline reasoning. Agentverse~\cite{agentverse} and CAPO~\cite{capo} enable iterative proposal, evaluation, and refinement of plans, supporting complex task decomposition. \textcolor{black}{More recently, Yuksel et al.~\cite{multiai} introduce an autonomous multi-agent refinement loop in which agents collectively propose and validate system-level modifications, enabling end-to-end self-optimization.} Debate-oriented frameworks such as SMoA~\cite{smoa} and MAD~\cite{MAD} further enhance decision quality by coordinating multiple agents under a judge or selector.


\textcolor{black}{Although parameter-free optimization methods primarily operate at inference time, recent research increasingly combines them with parameter-based approaches to achieve more stable and higher-capacity agent behavior. A growing body of recent work suggests that effective agent optimization increasingly benefits from combining parameter-based and parameter-free strategies, as each addresses complementary aspects of agent capability. For instance, Retrospex~\cite{retrospex} incorporates supervised fine-tuning with value-guided weighting during retrospection, SwiftSage~\cite{swiftsage} alternates between a fine-tuned model and prompt-enhanced reasoning, and NLRL~\cite{nlrl} expresses reinforcement signals in natural language that can subsequently be distilled through fine-tuning. These examples indicate that modern LLM-based agent systems often integrate both paradigms to achieve more reliable and adaptable behavior than either approach alone.}
\color{black}\section{Open-source Frameworks and Stacks}\label{sec5}

\subsection{Background and Introduction}
LLM-based agents, capable of autonomous perception, reasoning, decision-making, and action, have rapidly become a core paradigm for building complex AI applications. As these agents integrate natural-language interfaces, external tools, and environment interaction, their development requires standardized and reusable infrastructure. To support this need, open-source frameworks such as LangChain\footnote{\url{https://github.com/langchain-ai/langchain}}
 and AutoGen~\cite{autogen} offer modular abstractions for planning, memory management, tool invocation, and execution control, substantially reducing engineering complexity and serving as the foundation for many recent agent systems and optimization methods.

Despite differences in design, most LLM agent frameworks share a common set of core components. \textbf{\emph{Planning}} decomposes high-level goals into actionable substeps. \textbf{\emph{Tool use}} enables agents to invoke external APIs or functions to access information or perform operations beyond language generation. \textbf{\emph{Memory}} modules store short-term context and long-term experiential knowledge, allowing decisions to incorporate prior interactions. \textbf{\emph{Reflection}} mechanisms help agents evaluate outputs, identify errors, and refine future actions. Together, these components form a unified architectural foundation for building robust and adaptable LLM-based agents.

\subsection{Common Open-source Agent Frameworks}
We summarize several widely used and actively maintained open-source frameworks for building LLM-based agents in Table~\ref{tab:agent_frameworks}. The selection excludes low-code platforms such as Coze\footnote{\url{https://github.com/coze-dev/coze-studio}}, as well as frameworks that have been discontinued or are no longer actively maintained.

Among these frameworks, orchestration-centered systems such as LangChain and its workflow extension LangGraph emphasize modular composition and graph-structured control, enabling developers to chain LLM calls, integrate external tools, and manage multi-step workflows with fine-grained state tracking. LlamaIndex~\cite{Liu_LlamaIndex_2022} follows a similar orchestration philosophy but is distinguished by flexible data interfaces and a retrieval-centric design suited for applications that rely on structured or unstructured knowledge sources.
In contrast, collaboration-centered frameworks including AutoGen~\cite{autogen}, CrewAI~\cite{crewAI}, Camel-AI~\cite{camel}, and AgentScope~\cite{gao2024agentscope} focus on role-based agent interaction and coordinated multi-step problem solving. AutoGen provides an actor-style architecture for multi-agent conversation, tool execution, and iterative debugging, whereas CrewAI emphasizes ease of defining team-like roles and workflows. Camel-AI and AgentScope extend this pattern through specialized agent communication and scalable multi-agent environments. Semantic Kernel~\cite{SemanticKernel} offers a lightweight, multi-language runtime that integrates AI services into existing software ecosystems while supporting both single-agent and multi-agent patterns.
Overall, orchestration-oriented frameworks provide stronger support for workflow representation and tool or data integration, whereas collaboration-oriented frameworks offer greater diversity in multi-agent interaction patterns and role modeling.

\begin{table}[t]
\centering
\caption{Representative open-source frameworks and stacks for building LLM-based agents.}
\label{tab:agent_frameworks}

\renewcommand{\arraystretch}{1.15}
\scriptsize
\begin{tabularx}{\linewidth}{l X X X c}
\hline
\textbf{Framework} & \textbf{Core Capabilities} & \textbf{Strengths} & \textbf{Limitations} & \textbf{Code} \\
\hline
\centering \textbf{LangGraph} 
& Modular and graph-based orchestration 
& Visual workflow design; fine-grained control 
& Structural complexity; steep learning curve 
& \href{https://github.com/langchain-ai/langchain}{Link} \\
\hline
\centering \textbf{AutoGen} 
& Multi-agent dialogue coordination 
& Flexible collaboration; complex problem handling 
& High debugging cost; resource-intensive 
& \href{https://github.com/microsoft/autogen}{Link} \\
\hline
\textbf{CrewAI} 
& Role–task workflow design 
& Team-oriented collaboration; lightweight usage 
& Limited scalability; evolving ecosystem 
& \href{https://github.com/crewAIInc/crewAI}{Link} \\
\hline
\textbf{Camel-AI} 
& Role-driven interaction protocols 
& Natural role-play; simple agent specification 
& Narrow task scope; moderate scalability 
& \href{https://github.com/camel-ai/camel}{Link} \\
\hline
\textbf{Semantic Kernel} 
& Lightweight AI orchestration 
& Strong system integration; high developer efficiency 
& Platform coupling; less mature ecosystem 
& \href{https://github.com/microsoft/semantic-kernel}{Link} \\
\hline
\textbf{LlamaIndex} 
& Data-centric retrieval-oriented pipelines 
& Strong retrieval modules; rich data connectors 
& Data dependence; limited autonomous behavior 
& \href{https://github.com/run-llama/llama_index}{Link} \\
\hline
\textbf{AgentScope} 
& Scalable multi-agent execution 
& Engineering-ready design; distributed operation 
& High resource demand; higher integration overhead 
& \href{https://github.com/agentscope-ai/agentscope}{Link} \\
\hline
\end{tabularx}
\end{table}

\color{black}
\vspace{-0.5em}
\textcolor{black}{\section{Evaluation and Datasets for LLM-Based Agents} \label{sec6}
Evaluating LLM-based agents requires both appropriate assessment methodologies and diverse datasets that reflect agent performance in various domains. In this section, we first summarizes common categories of agent evaluation approaches, and then reviews the benchmarks used to measure agent performance as well as the datasets designed for agent tuning and optimization.} 

\textcolor{black}{\subsection{Evaluation Methodologies}
Current evaluation methodologies can be broadly grouped into two paradigms~\cite{mohammadi2025evaluation}.}

\textcolor{black}{\textbf{(1) Human evaluation} remains necessary for tasks that are subjective or lack clear quantitative metrics. It is commonly used for assessing text quality, creativity, coherence, emotional intelligence, or specialized reasoning-scenarios where correctness cannot be fully captured by predefined rules. Human judgments reflect user preferences and provide nuanced assessments but are costly, difficult to scale, and prone to variability, limiting their use in large-scale or iterative agent evaluation.
}

\textcolor{black}{
\textbf{(2) Automated evaluation} has become the primary approach for LLM-based agents due to its scalability and reproducibility. \textbf{Static dataset evaluation} relies on curated benchmarks and predefined metrics (e.g., accuracy or execution success rate) and works well for objective tasks~\cite{mohammadi2025evaluation}, though it may overlook interactive behaviors. \textbf{LLM-as-a-judge methods} extend automation to subjective tasks by using stronger models to score or compare outputs~\cite{li2025generation}, enabling automated assessment for subjective or open-ended tasks. Recent multi-dimensional automated protocols integrate rule-based metrics, execution feedback, and model-based judgments within interactive environments, enabling more holistic assessment of reasoning quality, tool-use reliability, and multi-step decision-making.
}
\subsection{Datasets and Benchmarks for Evaluation}

\subsubsection{\textbf{General Domain Evaluation Tasks}}
We summarize common evaluation datasets categorized by general task domains in Table~\ref{evaluate datasets}, including programming, question answering (QA), multimodal tasks. Following AgentBank~\cite{agentbank}, we also use action space to distinguish tasks where actions like natural language and code are classified into continuous space.
 
\begin{table}[htbp]
\setlength{\abovecaptionskip}{0.1cm}
\caption{Summary of Commonly Used Datasets and Environments for Evaluation Tasks.}
\label{evaluate datasets}
\begin{center}
\resizebox{0.9\textwidth}{!}{
\begin{tabular}{c|c|c|c|c}
\hline
\textbf{Domain} & \textbf{Datasets/Environments} & \textbf{Task Type} & \textbf{Action Space} & \textbf{Data Link} \\
\hline
\multirow{6}{*}{\centering Math} & GSM8K~\cite{gsm8k} & Mathematical reasoning  & Continuous & \href{https://openai.com/index/solving-math-word-problems/}{\textcolor{blue}{Link}} \\
 & AsDIV~\cite{asdiv} &  Mathematical reasoning & Continuous & \href{https://github.com/chaochun/nlu-asdiv-dataset}{\textcolor{blue}{Link}} \\
 & SVAMP~\cite{svamp} &  Mathematical reasoning & Continuous & \href{https://github.com/arkilpatel/SVAMP}{\textcolor{blue}{Link}} \\
 & MATH~\cite{math} &  Mathematical reasoning & Continuous & \href{https://github.com/openai/grade-school-math}{\textcolor{blue}{Link}} \\
 & AIME &  Mathematical reasoning & Continuous & \href{https://www.kaggle.com/datasets/hemishveeraboina/aime-problem-set-1983-2024}{\textcolor{blue}{Link}} \\
\hline
\multirow{7}{*}{\centering QA} & HotpotQA~\cite{hotpotqa} & Open-domain QA & Continuous & \href{https://hotpotqa.github.io/}{\textcolor{blue}{Link}} \\
 & MMLU~\cite{mmlu} & Multiple choice QA & Continuous & \href{https://github.com/hendrycks/test}{\textcolor{blue}{Link}} \\
 & TruthfulQA~\cite{truthfulqa} & Multiple choice QA, Open-domain QA & Continuous & \href{https://github.com/sylinrl/TruthfulQA}{\textcolor{blue}{Link}} \\
 & PubMedQA~\cite{pubmedqa} & Domain-Specific QA & Continuous & \href{https://pubmedqa.github.io/}{\textcolor{blue}{Link}} \\
 & MuSiQue~\cite{musique} & Commonsense Reasoning & Continuous & \href{https://github.com/stonybrooknlp/musique}{\textcolor{blue}{Link}} \\
 & QASPER~\cite{qasper} & Domain-Specific QA & Continuous & \href{https://huggingface.co/datasets/allenai/qasper}{\textcolor{blue}{Link}} \\
 & ARC~\cite{arc} & Commonsense Reasoning & Continuous & \href{https://huggingface.co/datasets/allenai/ai2_arc}{\textcolor{blue}{Link}} \\
\hline
\multirow{6}{*}{\centering Code} & SWE-bench~\cite{swe=bench} & Code Generation, Bug Fixing & Continuous & \href{https://www.swebench.com/}{\textcolor{blue}{Link}} \\
 & HumanEval~\cite{humaneval} & Code Generation & Continuous & \href{https://github.com/openai/human-eval}{\textcolor{blue}{Link}} \\
 & LiveCodeBench~\cite{livecodebench} & Code Execution, Bug Fixing & Continuous & \href{https://livecodebench.github.io/}{\textcolor{blue}{Link}} \\
 & BIRD~\cite{bird} & Text-to-SQL Conversion & Continuous & \href{https://bird-bench.github.io/}{\textcolor{blue}{Link}} \\
 & InterCodeSQL~\cite{intercode} & Code Generation & Continuous & \href{https://intercode-benchmark.github.io/}{\textcolor{blue}{Link}} \\
  & MLE-Bench~\cite{mle-bench} & algorithm design & Continuous & \href{https://github.com/openai/mle-bench}{\textcolor{blue}{Link}} \\
\hline
\multirow{4}{*}{\centering Tool-use} & T-Eval~\cite{t-eval} & Tool-use & Continuous & \href{https://open-compass.github.io/T-Eval/}{\textcolor{blue}{Link}} \\
 & MINT-Bench~\cite{mint} & Tool-use & Continuous & \href{https://xwang.dev/mint-bench/}{\textcolor{blue}{Link}} \\
 & ToolEval~\cite{toolllm} & Tool-use & Continuous & \href{https://github.com/OpenBMB/ToolBench}{\textcolor{blue}{Link}} \\
 & MTU-Bench~\cite{mtu-bench} & Tool-use & Continuous & \href{https://github.com/MTU-Bench-Team/MTU-Bench.git}{\textcolor{blue}{Link}} \\

\hline
\multirow{4}{*}{\centering Web} & WebShop~\cite{webshop} & Web Navigation & Discrete & \href{https://webshop-pnlp.github.io/}{\textcolor{blue}{Link}} \\
 & WebArena~\cite{webarena} & Web Navigation & Discrete & \href{https://webarena.dev/}{\textcolor{blue}{Link}} \\
 & Mind2Web~\cite{mind2web} & Web Navigation & Discrete & \href{https://osu-nlp-group.github.io/Mind2Web/}{\textcolor{blue}{Link}} \\
 & MiniWoB++~\cite{miniwob} & Web Navigation & Discrete & \href{https://github.com/Farama-Foundation/miniwob-plusplus}{\textcolor{blue}{Link}} \\
\hline
\multirow{8}{*}{\centering Environment Interaction} & ScienceWorld~\cite{scienceworld} & Scientific Experiment & Discrete & \href{https://sciworld.apps.allenai.org/}{\textcolor{blue}{Link}} \\
 & ALFWorld~\cite{ALFWorld} & Embodied AI & Discrete & \href{https://alfworld.github.io/}{\textcolor{blue}{Link}} \\
 & TDW-MAT~\cite{tdw-mat} & Embodied AI & Discrete & \href{https://github.com/UMass-Foundation-Model/Co-LLM-Agents/}{\textcolor{blue}{Link}} \\
 & C-WAH~\cite{tdw-mat} & Embodied AI & Discrete & \href{https://github.com/UMass-Foundation-Model/Co-LLM-Agents/}{\textcolor{blue}{Link}} \\
 & ALFRED~\cite{alfred} & Embodied AI & Discrete & \href{https://askforalfred.com/}{\textcolor{blue}{Link}} \\
 & RLCard~\cite{rlcard}& Game & Discrete & \href{https://rlcard.org/games.html}{\textcolor{blue}{Link}} \\
 & Breakthrough~\cite{openspiel} & Game & Discrete & \href{https://github.com/google-deepmind/open_spiel?tab=readme-ov-file}{\textcolor{blue}{Link}} \\
\hline
\multirow{5}{*}{\centering Multimodal} & VQA v2.0~\cite{vqa} & Visual QA & Continuous & \href{https://visualqa.org/}{\textcolor{blue}{Link}} \\
 & A-OKVQA~\cite{a-okvqa} & Visual QA & Continuous & \href{https://github.com/allenai/aokvqa}{\textcolor{blue}{Link}} \\
& ScienceQA (IMG)~\cite{scienceqa} & Visual QA & Continuous & \href{https://scienceqa.github.io/}{\textcolor{blue}{Link}} \\
 & EgoSchema~\cite{egoschema} & Video Understanding, Visual QA & Continuous & \href{https://egoschema.github.io/}{\textcolor{blue}{Link}} \\
 & NExT-QA~\cite{nextqa} & Video Understanding, Visual QA & Continuous & \href{https://github.com/doc-doc/NExT-QA}{\textcolor{blue}{Link}} \\
\hline

\end{tabular}
}
\end{center}
\end{table}

\textbf{Math:}
Mathematical reasoning tasks assess an agent’s ability to perform multi-step inference and solve quantitative problems. \textcolor{black}{Common evaluation metrics include exact-match (EM) accuracy and Pass@k, which measure the correctness of final generated answers under deterministic or sampled decoding. Representative datasets include GSM8K~\cite{gsm8k} for arithmetic reasoning, SVAMP~\cite{svamp} for robustness to linguistic variations, and MATH~\cite{math} or AIME for evaluating advanced competition-level problem-solving.}

\textbf{QA:}
\textcolor{black}{Question answering tasks assess an agent’s ability to comprehend text, retrieve relevant information, and perform single- or multi-step reasoning to produce accurate answers. Common evaluation metrics include EM and F1 score, which measure correctness and overlap with gold-standard answers; multi-choice datasets typically use accuracy.} Representative benchmarks include HotpotQA~\cite{hotpotqa} and MuSiQue~\cite{musique} for evaluating multi-hop reasoning across documents, and MMLU~\cite{mmlu} for broad knowledge coverage across diverse academic subjects. Domain-specific datasets such as PubMedQA~\cite{pubmedqa} assess reasoning over biomedical literature, while ARC~\cite{arc} focuses on scientific reasoning at the grade-school level.

\textbf{Code:}
\textcolor{black}{Code-grounded tasks evaluate an agent’s ability to understand specifications, generate executable programs, fix bugs, and reason over large codebases. Common evaluation metrics include pass@k for function-level code synthesis, execution success rate for verifying correctness against test cases, and runtime or efficiency of programs. For end-to-end algorithmic challenges such as MLE-Bench~\cite{mle-bench}, evaluation focuses on machine learning task performance (e.g. acc, loss), reflecting an agent’s capability in algorithm design.}
HumanEval~\cite{humaneval} provides a standard setting for Python function synthesis, while SWE-bench~\cite{swe=bench} targets real-world software engineering tasks that require generating repository-grounded patches. LiveCodeBench~\cite{livecodebench} extends evaluation to competitive programming problems, and database-oriented benchmarks such as BIRD~\cite{bird} and InterCodeSQL~\cite{intercode} measure SQL generation and refinement through execution feedback.

\textbf{Tool-use:}
\textcolor{black}{Tool-use tasks assess an agent’s ability to plan and execute interactions with external tools or APIs, requiring accurate tool selection, parameter construction, and integration of tool outputs into multi-step reasoning. Common evaluation metrics include tool selection accuracy, tool-call success rate, and tool amount or order accuracy in multi-turn tool-use settings.  
Representative datasets include T-Eval~\cite{t-eval} for fine-grained diagnostics of planning and tool invocation, ToolEval~\cite{toolllm} for automated functional scoring, and MTU-Bench~\cite{mtu-bench} for multi-granularity scenarios with prediction-based metrics such as tool selection and parameter accuracy. MINT-Bench~\cite{mint} further evaluates multi-turn tool interactions driven by natural-language feedback.}

\textbf{Web:}
\textcolor{black}{Web-based tasks evaluate an agent’s ability to navigate websites, understand structured and unstructured content, and complete goal-directed interactions. Common evaluation metrics include task success rate, navigation efficiency (e.g., number of steps), and action accuracy for page-level operations.} Representative benchmarks include WebShop~\cite{webshop} for instruction-following in e-commerce settings and WebArena~\cite{webarena} for interacting with full-featured web applications. Larger-scale datasets such as Mind2Web~\cite{mind2web} and MiniWoB++~\cite{miniwob} further provide diverse real-world or synthetic environments for evaluating multi-step web navigation behaviors.

\textbf{Environment Interaction:} 
Environment interaction tasks evaluate an agent’s ability to perform multi-step decision-making and adapt to feedback within dynamic physical or virtual environments. \textcolor{black}{Common evaluation metrics include task success rate, average score and reward or win rate in game-based settings.} Representative benchmarks include ScienceWorld~\cite{scienceworld} for text-based scientific experimentation and embodied environments such as ALFWorld~\cite{ALFWorld} and ALFRED~\cite{alfred}, which require language-guided object manipulation and navigation. Game-oriented platforms like RLCard~\cite{rlcard} and OpenSpiel~\cite{openspiel} further assess strategic planning and multi-agent decision-making across competitive and cooperative tasks.

\textbf{Multimodal:}
In multimodal tasks, agents integrate information across images, text, and videos. \textcolor{black}{Common evaluation metrics include answer accuracy, VQA accuracy, multiple-choice accuracy for image–text reasoning, and temporal reasoning accuracy for video-based tasks. VQA-V2~\cite{vqa} tests core visual question answering abilities, while A-OKVQA~\cite{a-okvqa} and ScienceQA (IMG)~\cite{scienceqa} introduce knowledge-intensive visual reasoning challenges. Video benchmarks ~\cite{egoschema,nextqa} further assess temporal, causal, and activity-level understanding in dynamic visual scenes.}

\subsubsection{\textbf{Multi-task Benchmarks}}
\textcolor{black}{To evaluate agents’ ability to generalize across diverse domains, multi-task benchmarks integrate heterogeneous task types into unified evaluation suites, providing a comprehensive assessment of performance.
Datasets such as AgentBench and AgentEval focus on programming, web navigation, and tool-augmented decision-making, often using unified formats such as ReAct to consolidate reasoning and action traces.
Benchmarks including Just-Eval, StreamBench, and AgentBoard extend evaluation to broader scenarios—text reasoning, embodied interaction, and game environments—aggregating diverse settings to assess generalization across varied modalities and task structures. GAIA introduces a realistic assistant-oriented paradigm requiring reasoning, multimodal understanding, web interaction, and tool usage, offering a rigorous benchmark for progress toward general-purpose AI assistants.
Table~\ref{multi_task_benchmarks} summarizes representative benchmarks, including their coverage, domain types, sample sizes, and resource links.} 

\begin{table}[htbp]
\setlength{\abovecaptionskip}{0.1cm}
\caption{Multi-Task Agent Evaluation Benchmarks. We use \ding{172}, \ding{173}, \ding{174}, etc., to represent the covered task domains: \ding{172} QA, \ding{173} Math, \ding{174} Web, \ding{175} Embodied AI, \ding{176} Code, \ding{177} Tools, \ding{178} Game, \ding{179} Writing, \ding{180} Role-playing.}
\begin{center}
\resizebox{0.9\textwidth}{!}{%
\begin{tabular}{c|c|c|c|c|c}
\hline
\textbf{Benchmark} & \textbf{\# Datasets} & \textbf{\# Task Types} & \textbf{Test Size} & \textbf{Covered Domains} & \textbf{Data Link} \\
\hline
AgentBench~\cite{agentbench} & 8 & 3 & 1,091 & \ding{174} \ding{176} \ding{178} & \href{https://github.com/THUDM/AgentBench}{\textcolor{blue}{Link}} \\
\hline
AgentEval~\cite{agentgym} & 14 & 5 & 1,160 & \ding{174} \ding{175} \ding{176} \ding{177} \ding{178}& \href{https://huggingface.co/datasets/AgentGym/AgentEval}{\textcolor{blue}{Link}} \\
\hline
Just-Eval~\cite{justeval}& 9 & 7 & 1,000 & \ding{172} \ding{173} \ding{176} \ding{179} \ding{180}& \href{https://github.com/Re-Align/just-eval}{\textcolor{blue}{Link}} \\
\hline
StreamBench~\cite{streambench} & 7 & 5 & 9,702 & \ding{172} \ding{176} \ding{177}&   \href{https://stream-bench.github.io/}{\textcolor{blue}{Link}} \\
\hline
AgentBoard~\cite{agentboard} & 9 & 4 & 1,013 & \ding{174} \ding{175} \ding{177} \ding{178} & \href{https://hkust-nlp.github.io/agentboard/}{\textcolor{blue}{Link}} \\
\hline
GAIA~\cite{mialon2023gaia} & - & 5 & 466 & \ding{172} \ding{174} \ding{176} \ding{177} & \href{https://huggingface.co/gaia-benchmark}{\textcolor{blue}{Link}} \\

\hline
\end{tabular}%
}
\label{multi_task_benchmarks}
\end{center}
\end{table}


\subsection{Datasets for Agent Tuning}
Fine-tuning datasets provide trajectories that equip LLM-based agents with task-specific behaviors, enabling them to acquire skills such as tool use, web interaction, programming, mathematical reasoning, and embodied decision-making. Recent tuning corpora such as AgentInstruct, AgentBank, Agent-FLAN, AgentOhana, FireAct, ToRA-CORPUS, AgentTraj, and SMART-Trajectory—integrate trajectories collected or synthesized from diverse sources including web environments, embodied platforms, SQL querying tasks, scientific reasoning datasets, and multi-turn ReAct-style interactions. These datasets typically combine curated demonstrations with filtered trajectories from large models, standardize multi-turn formats, and cover heterogeneous action spaces, thereby supporting scalable behavioral cloning and instruction-following fine-tuning across a wide range of agent scenarios. Table~\ref{finetune data} summarizes their task coverage, trajectory volume, and data sources.

\begin{table}[htbp]
\setlength{\abovecaptionskip}{0.1cm}
\caption{Datasets for Agent Tuning. We use \ding{172}, \ding{173}, \ding{174}, etc., to represent the task domains as follows: \ding{172} QA (Commonsense Reasoning and Open-domain QA), \ding{173} Math, \ding{174} Web, \ding{175} Embodied AI, \ding{176} Code, \ding{177} Tools, \ding{178} Game, \ding{179} Fact Verification, \ding{180} Dialogue, \ding{181} General Instructions.}

\begin{center}
\resizebox{0.9\textwidth}{!}{%
\begin{tabular}{c|c|c|c|c}
\hline
\textbf{Dataset} & \textbf{\# Tasks} & \textbf{\# Filtered Trajectories} & \textbf{Covered Domains} & \textbf{Data Link} \\
\hline
AgentInstruct~\cite{agenttuning} & 6 & 1,866 & \ding{174} \ding{175} \ding{176}  & \href{https://huggingface.co/datasets/THUDM/AgentInstruct}{\textcolor{blue}{Link}} \\
AgentBank~\cite{agentbank} & 16 & 51,287 & \ding{172} \ding{173} \ding{174}  \ding{175}  \ding{176} & \href{https://huggingface.co/datasets/Solaris99/AgentBank}{\textcolor{blue}{Link}} \\
Agent-FLAN~\cite{agentflan} & 7 & 24,703 & \ding{174} \ding{176} \ding{177}  & \href{https://huggingface.co/datasets/internlm/Agent-FLAN}{\textcolor{blue}{Link}} \\
AgentOhana~\cite{agentohana} & 10 & 42,600 & \ding{174} \ding{176} \ding{177}  & \href{https://github.com/SalesforceAIResearch/xLAM}{\textcolor{blue}{Link}} \\
FireAct~\cite{fireact} & 3 & 1,344 & \ding{172}  & \href{https://fireact-agent.github.io/}{\textcolor{blue}{Link}} \\
ToRA-CORPUS~\cite{tora} & 2 & 15,538 & \ding{173}  & \href{https://github.com/microsoft/ToRA}{\textcolor{blue}{Link}} \\
AgentTraj/AgentTraj-L~\cite{agentgym} & 14 & 6,130/14,485 & \ding{174} \ding{175} \ding{176} \ding{177} \ding{178}  & \href{https://huggingface.co/datasets/AgentGym/AgentTraj-L}{\textcolor{blue}{Link}} \\
SMART-Trajectory~\cite{smart} & 17 & 142,507 & \ding{172} \ding{179} \ding{180} \ding{181} & \href{https://github.com/yueshengbin/SMART}{\textcolor{blue}{Link}} \\
\hline
\end{tabular}
}
\end{center}
\label{finetune data}
\end{table}

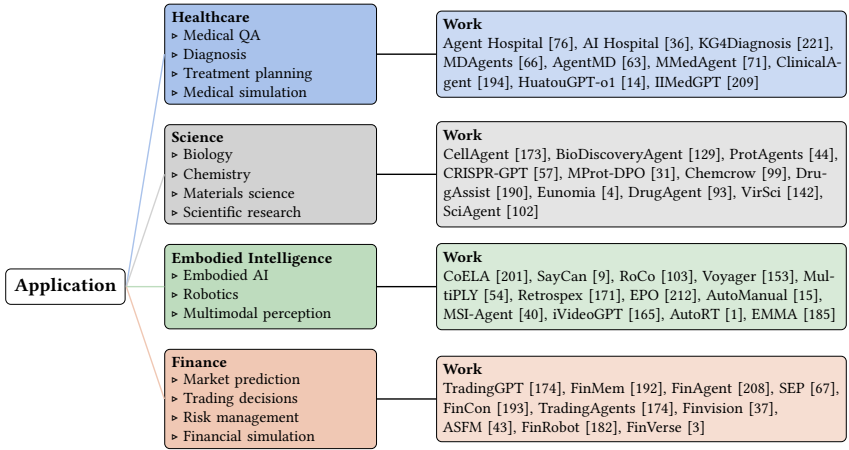
\begin{figure}[htbp]
  \centering
  \resizebox{0.8\linewidth}{!}{%
\begin{tikzpicture}[node distance=4mm and 12mm,]

\node[cat=health] (health) at (1,4) {%
  \textbf{Healthcare}\\[-1pt]
  $\triangleright$ Medical QA\\
  $\triangleright$ Diagnosis\\
  $\triangleright$ Treatment planning\\
  $\triangleright$ Medical simulation};

\node[cat=sci, below=of health] (science) {%
  \textbf{Science}\\[-1pt]
  $\triangleright$ Biology\\
  $\triangleright$ Chemistry\\
  $\triangleright$ Materials science\\
  $\triangleright$ Scientific research};

\node[cat=emb, below=of science] (emb) {%
  \textbf{Embodied Intelligence}\\[-1pt]
  $\triangleright$ Embodied AI\\
  $\triangleright$ Robotics\\
  $\triangleright$ Multimodal perception};

\node[cat=fin, below=of emb] (fin) {%
  \textbf{Finance}\\[-1pt]
  $\triangleright$ Market prediction\\
  $\triangleright$ Trading decisions\\
  $\triangleright$ Risk management\\
  $\triangleright$ Financial simulation};


\node[root] (app) at ($(emb.west)+(-20mm,0)$) {Application};
\coordinate (appE) at ($(app.east)+(0.5pt,0)$);  
\node[work=health, right=of health] (healthW) {%
  \textbf{Work}\\
Agent Hospital~\cite{agenthospital}, AI Hospital~\cite{aihospital}, KG4Diagnosis~\cite{kg4diagnosis},
  MDAgents~\cite{mdagents}, AgentMD~\cite{agentmd},
  MMedAgent~\cite{mmedagent}, ClinicalAgent~\cite{clinicalagent}, HuatouGPT‑o1~\cite{huatuogpt}, IIMedGPT~\cite{iimedgpt}};

\node[work=sci, right=of science] (scienceW) {%
  \textbf{Work}\\
  CellAgent~\cite{cellagent}, BioDiscoveryAgent~\cite{biodiscoveryagent}, ProtAgents~\cite{ghafarollahi2024protagents},
  CRISPR‑GPT~\cite{crispr}, MProt‑DPO~\cite{mprot-}, Chemcrow~\cite{ChemCrow},
  DrugAssist~\cite{drugassist}, Eunomia~\cite{Eunomia}, DrugAgent~\cite{drugagent}, VirSci~\cite{VirSci}, SciAgent~\cite{sciagent}};

\node[work=emb, right=of emb] (embW) {%
  \textbf{Work}\\
  CoELA~\cite{tdw-mat}, SayCan~\cite{saycan}, RoCo~\cite{roco}, Voyager~\cite{voyager},
  MultiPLY~\cite{multiply}, Retrospex~\cite{retrospex}, EPO~\cite{epo}, AutoManual~\cite{automanual},
  MSI‑Agent~\cite{msi-AGENT}, iVideoGPT~\cite{ivideogpt}, AutoRT~\cite{autort}, EMMA~\cite{emma}};

\node[work=fin, right=of fin] (finW) {%
  \textbf{Work}\\
  TradingGPT~\cite{tradingagents}, FinMem~\cite{finmem}, FinAgent~\cite{FINEAGENT}, SEP~\cite{sep}, FinCon~\cite{fincon}, TradingAgents~\cite{tradingagents}, Finvision~\cite{finvision}, ASFM~\cite{ASFM}, FinRobot~\cite{finrobot}, FinVerse~\cite{finverse}};


\draw[thick, draw=health] (appE) -- (health.west);   
\draw[thick, draw=sci]    (appE) -- (science.west);  
\draw[thick, draw=emb]    (appE) -- (emb.west);      
\draw[thick, draw=fin]    (appE) -- (fin.west);      

\foreach \src/\dst in {health/healthW,
                       science/scienceW,
                       emb/embW,
                       fin/finW}
  \draw[thick] (\src.east) -- (\dst.west);

\end{tikzpicture}}
  \caption{The applications of LLM-based Agents.}
    \label{fig:application}
\end{figure}
\section{Application} \label{sec7}

LLM-based agents have been applied across diverse domains, showing their potential to address complex tasks and enhance productivity. This section provides an overview of representative applications in Healthcare, Science, Embodied Intelligence, and Finance, as illustrated in Figure~\ref{fig:application}.

\subsection{Healthcare}

LLM-based agents in healthcare generally follow a progressive workflow that spans medical question answering, diagnosis, treatment planning, and medical simulation, with early systems enhancing conversational accuracy through domain-specific fine-tuning~\cite{medpalm,disc-med}. Subsequent work extends these capabilities through simulation frameworks~\cite{agenthospital,aihospital}, which provide dynamic clinical environments for realistic physician–patient interactions, and through multi-agent systems such as KG4Diagnosis~\cite{kg4diagnosis} and MDAgents~\cite{mdagents}, which model collaboration between general practitioners and specialists to improve diagnostic reasoning. More recent methods~\cite{huatuogpt,iimedgpt} further combine fine-tuning with RL techniques such as PPO or DPO to strengthen long-horizon clinical reasoning and treatment optimization. Despite these advances, key challenges persist, including the risks posed by medical hallucinations, strict privacy and compliance constraints, and the limited context capacity of LLMs for processing full electronic medical records.

\subsection{Science}
LLM-based agents are increasingly used to support scientific research by integrating domain knowledge retrieval, hypothesis generation, and automated experiment design into unified workflows that accelerate biological and chemical discovery. Recent work in biology, such as CellAgent~\cite{cellagent} and BioDiscoveryAgent~\cite{biodiscoveryagent}, shows how multi-agent coordination and iterative self-optimization can automate tasks like single-cell data analysis and gene perturbation studies. Tool-integrated agents including ProtAgents~\cite{ghafarollahi2024protagents} and CRISPR-GPT~\cite{crispr} further leverage external scientific software to assist protein design and gene editing.
Emerging frameworks now aim to design more general AI scientist agents. AlphaEvolve~\cite{novikov2025alphaevolve} illustrates how code-evolving agents can autonomously discover more efficient algorithms and scientific constructs, while InternAgent~\cite{team2025internagent} introduces a closed-loop multi-agent system spanning idea generation, methodology design, and experimental execution.
However, scientific workflows remain challenging due to high computational costs, long knowledge-retrieval cycles, and the complexity of coordinating multi-stage experimental pipelines, highlighting the need for more efficient and scalable scientific agent systems.

\subsection{Embodied Intelligence}
In the fields of embodied intelligence and robotic systems, \textcolor{black}{LLM-based agents} autonomously interact with the physical world, and integrate multimodal inputs such as visual and auditory inputs to perform complex tasks. 
\textcolor{black}{
Early systems demonstrate how perception, memory, and planning modules can be combined to execute complex embodied tasks~\cite{tdw-mat,roco}, while RL-based approaches such as SayCan~\cite{saycan} ground LLM-driven reasoning in real-world robotic actions. In interactive simulation settings, agents like Voyager~\cite{voyager} illustrate how long-term memory and iterative skill acquisition support continual learning. More recent frameworks~\cite{multiply,retrospex,epo,automanual,msi-AGENT} refine action strategies through environmental feedback and structured sub-goal decomposition, and multimodal methods ~\cite{ivideogpt,autort,emma} integrate visual, textual, and action signals to improve task planning and execution. Despite these advances, embodied systems face persistent challenges, including uncertainty in real-world environments, safety and reliability concerns for physical actuation, and the difficulty of maintaining consistent performance over long-horizon tasks.}

\subsection{Finance}
LLM-based agents are increasingly adopted in financial applications ranging from market prediction and trading decisions to risk management and financial simulation, where they process heterogeneous financial signals and support data-driven reasoning. \textcolor{black}{A variety of agent architectures have been explored: models equipped with hierarchical memory and reflective evaluation~\cite{tradinggpt,finmem,FINEAGENT} aim to capture shifting market patterns, and trading agents that incorporate finance signals such as SEP~\cite{sep} adapt their strategies through iterative policy refinement. MAS~\cite{fincon,tradingagents,finvision} extend financial analysis by coordinating diverse roles, enabling analyst-style reasoning, risk monitoring, and collaborative planning. Meanwhile, cross-modal frameworks like ASFM~\cite{ASFM} and FinRobot~\cite{finrobot} fuse textual, numerical, and visual financial information to support decisions in complex market settings, and systems such as FinVerse~\cite{finverse} augment analysis with code-execution capabilities for fine-grained financial computation. Applying these agents in practice requires careful consideration of financial data sensitivity, the fast-changing nature of market knowledge, and the demand for transparent and trustworthy decision-making in high-stakes environments.}

\section{Challenges \& Future Directions}\label{sec8}
We outline the key challenges faced in current state of LLM-based agent optimization and explore potential future research directions.

\subsection{Algorithm Adaptability and Efficiency}
A key problem in LLM-based agent optimization is balancing algorithmic efficiency with task-specific adaptability. Current methods, such as RL and fine-tuning, often struggle with sparse rewards, large action spaces or overfitting to specific data. Algorithms like PPO are effective or policy optimization but computationally expensive, while DPO simplifies  training but is primarily suited for single-step optimization, limiting its effectiveness for multi-step interactive tasks. Moreover, many existing methods fine-tune LLM-based agents using agent trajectory data for specific domains, which results in poor performance on unseen domains, particularly when there is a significant mismatch between training and real-world data distributions.
 
Exploring hybrid approaches that combine RL and fine-tuning offers a potential pathway to improving both adaptability and efficiency. Optimizing reward design and integrating efficiency-focused algorithms will also be crucial for reducing computational costs while maintaining high adaptability. Additionally, techniques like distribution alignment and domain adaptation can enable agents to transfer knowledge more effectively between tasks, even in dynamic environments.



\subsection{Standardized Evaluation Metrics}
The evaluation of LLM-based agents currently lacks standardized metrics, making fair performance comparison across tasks challenging. Different agents operate in varied environments, such as mathematical reasoning, web navigation, and embodied AI, each relying on distinct evaluation criteria. This variability makes it difficult to compare optimization methods fairly and assess their generalizability.
Additionally, existing metrics primarily assess task completion rather than extent of optimization, making it hard to quantify the effectiveness of different enhancement techniques. 

Establishing standardized evaluation frameworks provides a promising direction for facilitating fair comparisons across diverse LLM-based agent tasks, such as developing unified benchmarks that assess optimization effectiveness beyond task completion. Furthermore, creating proper metrics that track stepwise optimization progress and integrating preference-based evaluation signals could provide a more comprehensive assessment.

\vspace{-0.5 em}
\subsection{Cost and Efficiency Constraints in LLM-Based Agents}
LLM-based agents face substantial cost and efficiency constraints when deployed in long-horizon or tool-augmented settings. Achieving strong performance frequently depends on large-scale or proprietary reasoning models, whose API usage and on-premise deployment impose significant financial and computational overhead. Multi-turn interaction and extended reasoning chains further amplify latency and token consumption, making many real-world workflows difficult to scale. Although advanced reasoning models can enhance decision quality, their integration typically increases inference depth and cumulative token cost, limiting the practicality of deploying LLM-based agents in demanding operational environments.

Efforts to alleviate these constraints focus on reducing redundant inference and lowering the cost of environment construction. Techniques such as context optimization~\cite{ace}, adaptive memory mechanisms, and inference-efficient prompt engineering offer promising pathways for limiting token usage in multi-step interactions. In parallel, scalable training environments—particularly those that can be automatically generated or locally deployed—help reduce development overhead and enable more cost-effective agent training. Advancing these directions will be critical for making LLM-based agents viable at scale in long-horizon and resource-intensive scenarios.
\vspace{-0.5 em}
\subsection{Safety and Robustness of Autonomous Agents} 
LLM-based agents are becoming increasingly capable across planning, tool interaction, and multi-step decision-making. However, their expanded autonomy also exposes them to safety risks during real-world deployment. Agents remain vulnerable to prompt injection and jailbreak attacks, where crafted inputs can override intended behaviors or bypass system constraints. In tool use setting, incorrect or manipulated intermediate outputs may propagate into external actions, leading to unintended or unsafe executions. Moreover, when operating across diverse or shifting environments, agents often exhibit limited generalization, resulting in inconsistent or unpredictable behavior.

Mitigating these risks requires defense mechanisms at both system and model levels. Recent work~\cite{autodefense} introduces multi-agent defense frameworks to filter harmful responses and counter jailbreak-style prompts across models. System-level strategies including input–output filtering, fact-checking pipelines, multi-model consensus, and human-in-the-loop validation help constrain unsafe behaviors in multi-step interactions. At the model level, alignment-driven training and safety-aware optimization remain essential for producing reliable and policy-compliant outputs.

\subsection{LLM-Based Multi-Agent System Optimization}
LLM-based multi-agent optimization remains underexplored, with most approaches improving single agent while leaving the collective behavior of the system largely unaddressed. Existing multi-agent strategies typically rely on frozen LLMs and prompt-based workflow, limiting joint optimization and leading to inconsistent communication and conflicting objectives.
Recent empirical studies, such as MAST~\cite{cemri2025}, further show that many MAS failures stem not from model limitations but from structural issues—insufficient role specification, inter-agent misalignment, and weak validation or termination mechanisms. These observations highlight that advancing multi-agent optimization requires both parameter-level improvements and more reliable coordination structures.

Developing robust multi-agent parameter optimization methods is essential for agent systems~\cite{smart}. Joint parameter tuning and effective information exchange should be further explored to align agent objectives and maximize system performance. Moreover, reward-sharing mechanisms and hierarchical decision-making could further enable LLM-based MAS to operate effectively in dynamic and complex environments.
\section{Conclusion}
This survey provides a comprehensive overview of the optimization methods for LLM-based agents, categorizing them into parameter-driven and parameter-free approaches. We focus on parameter-driven optimization, encompassing conventional fine-tuning, RL-based methods, and hybrid strategies. We first detail the entire workflow from data construction to fine-tuning in the conventional fine-tuning part. Subsequently, we outline RL-based optimization, covering both reward function optimization and preference alignment methods. Then, we introduce parameter-free approaches, including historical experience integration, feedback mechanisms, tool usage, retrieval, and multi-agent collaboration. 
Furthermore, we summarize datasets widely utilized to evaluate or fine-tune agents and introduce the diverse real-world applications of LLM-based agents.
Finally, we identify several key challenges and propose future research directions aimed at driving the development of more capable and intelligent LLM-based agents.

\bibliographystyle{ACM-Reference-Format}
\bibliography{sample-base}

\end{document}